\documentclass[onecolumn]{robbyant}           %

\usepackage{longtable} 
\usepackage{makecell}
\usepackage{booktabs}
\usepackage{multirow}
\usepackage[table]{xcolor}
\usepackage{tabularx}

\metadata[Website]{\url{https://technology.robbyant.com/lingbot-vla-v2}}
\metadata[Github]{\url{https://github.com/robbyant/lingbot-vla-v2}}
\metadata[Checkpoints]{\url{https://huggingface.co/collections/robbyant/lingbot-vla-v2}}
\newcommand{\methodname}{LingBot-VLA}
\newcommand{\methodold}{\texttt{\methodname}\xspace}
\newcommand{\method}{\texttt{\methodname~2.0}\xspace}
\newcommand{\robotnum}{{20}\xspace}
\newcommand{\dinomodel}{DINO-Video\xspace}

\title{From Foundation to Application: \\[4pt] Improving VLA Models in Practice}

\author{
\begin{center}
    Wei Wu$^{*}$,
    Fangjing Wang$^{*}$,
    Fan Lu,
    He Sun,
    Shi Liu,
    Yunnan Wang,
    Yibin Yan,
    Yong Wang,
    \\[2pt]
    Shuailei Ma,
    Xinyang Wang,
    Yibin Liu,
    Shuai Yang,
    Tianxiang Zhou,
    Kejia Zhang,
    Lei Zhou,
    Cheng Su,
    \\[2pt] 
    Nan Xue,
    Bin Tan,
    Han Zhang,
    Youchao Zhang,
    Fei Liao,
    Xing Zhu,
    Yujun Shen,
    Kecheng Zheng$^{\dagger}$
    \\[12pt]
    {$^{*}$Equal Contribution} \qquad
    $^{\dagger}$Project Lead
\end{center}
}

\begin{document}

\vspace{-30pt}
\abstract{
Despite recent progress of VLA foundation models, the disparity between laboratory conditions and real-world applications continues to impede their practical implementation.
To bridge this gap, we present \method, which advances \methodold through improvements in three functional domains.
(1) \textbf{\textit{Generalization}} across tasks and embodiments.
Compared to the previous version, we revamp the data processing pipeline and curate around 60,000 hours of data for pretraining, including 50,000 hours of robot trajectories spanning \robotnum robot configurations and 10,000 hours of egocentric human videos.
(2) \textbf{\textit{Expanded action space}} in addition to dual-arm hardware platforms.
In particular, our system accommodates degrees of freedom for the heads, waists, mobile bases, and dexterous hands, thereby empowering the robots to tackle more complex tasks in practical scenarios.
(3) \textbf{\textit{Predictive dynamics modeling}} for improved temporal reasoning.
Specifically, we formulate future prediction as a proxy task, facilitated by a video representation model for semantic priors and a depth estimation model for geometric cues.
Evaluations on the GM-100 benchmark, conducted in a generalist setting, validate the beneficial impact of these proposed modifications.
Furthermore, benefiting from the expanded pretraining data that covers whole-body degrees of freedom, LingBot-VLA-2.0 demonstrates strong cross-embodiment long-horizon mobile manipulation capability across the two robotic platforms.

}

\maketitle

\justifying
\begin{figure}[t]
\centering
\includegraphics[width=\linewidth]{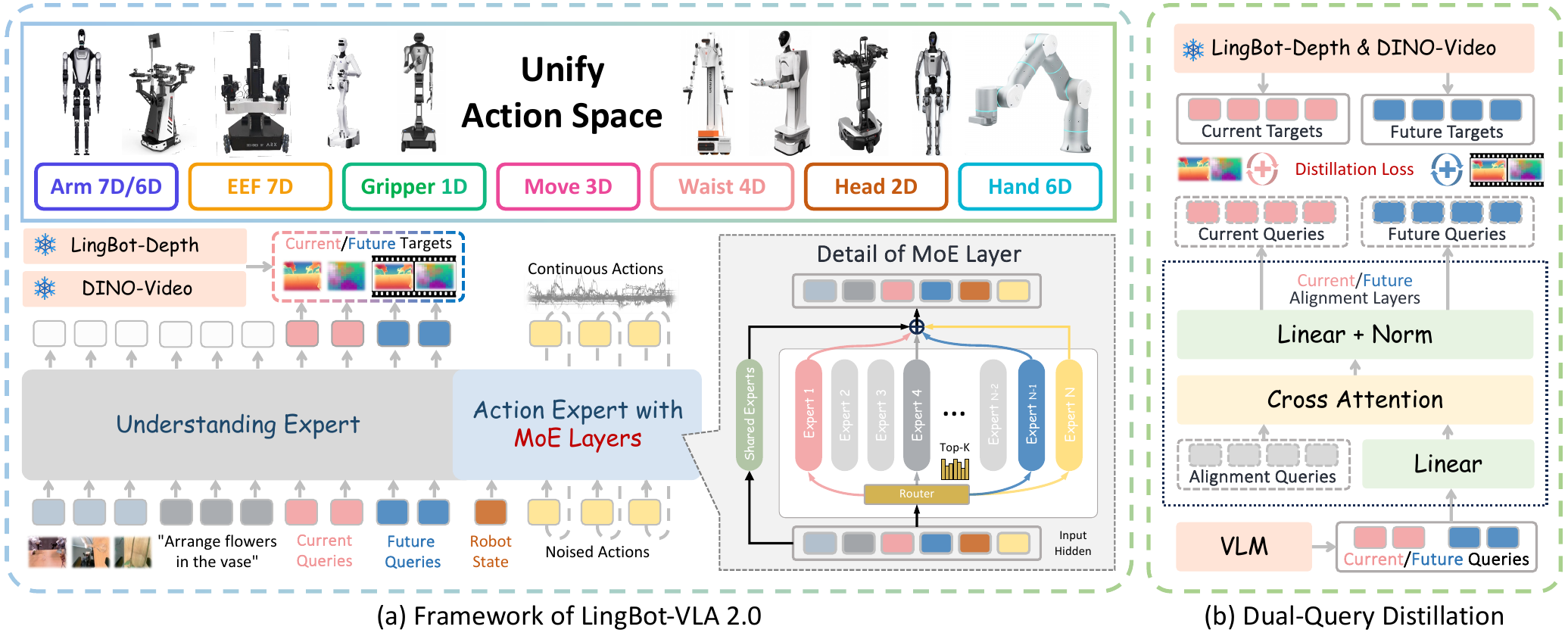}
\caption{\textbf{Overview of \method}. We revamp the data processing pipeline and curate 60,000 hours of pretraining data, including 50,000 hours of robot trajectories across 20 robot configurations and 10,000 hours of egocentric human videos. Moreover, our model supports degrees of freedom for the head, waist, mobile base, and dexterous hands, enabling robots to handle more complex real-world tasks. We formulate future prediction as a proxy task, leveraging a video representation model for semantic priors and a depth estimation model for geometric cues.}
\label{fig:teaser}
\end{figure}

\section{Introduction}\label{sec:intro}

Vision-language-action (VLA) models~\cite{pi_0,pi_0d5,gr00tN1,openvla} have recently emerged as a promising paradigm for building generalist robot policies. 
A key advantage of this paradigm is that pretrained vision-language models provide rich multimodal alignment and semantic representations, enabling VLA models to better understand complex scenes and generalize across diverse tasks.
Beyond such model-level priors, recent advances~\cite{pi_0d5,generalist2026gen1} further show that scaling up robot data in both quantity and diversity can substantially improve the capability of VLA systems.
Together, these developments have established VLA as a compelling foundation for robot learning.

However, despite this rapid progress, a substantial gap remains between laboratory benchmarks and real-world deployment. 
In practice, robots are expected to operate under broader embodiment diversity, richer action spaces, and more dynamic environments than those considered in many existing VLA settings. 
First, generalization in practice is not only about transferring across tasks, but also about handling heterogeneous robot configurations and data sources. 
Another point is that many real-world platforms involve substantially more degrees of freedom than standard dual-arm manipulation setups, including head movement, waist, mobile-base control, and dexterous hands. 
Subsequently, real-world execution often requires anticipating future scene evolution and action consequences, rather than reacting only to current observations. 
These challenges collectively limit the practical utility of current VLA foundation models.

A growing body of work has started to address these issues from different perspectives. Some approaches~\cite{beingH05,generalist2026gen1,lingbotvla} scale robot pretraining data to improve cross-task and cross-embodiment robustness.
Others incorporate embodiment-aware architectural designs~\cite{g0_5} to better handle heterogeneous robots. 
Another line of research augments VLA models with latent action model~\cite{dexWorldModel,lda1b} to improve decision making in dynamic environments. Meanwhile, recent system-oriented efforts have emphasized that practical VLA deployment depends not only on model scale, but also on data quality, action coverage, and training objectives that better align pretraining with downstream execution. These trends indicate that improving VLA models in practice requires a coordinated treatment of data, embodiment, and predictive capability.

In this work, we build on this perspective and present \method, an improved version of \methodold aimed at bridging foundation-level VLA capabilities with practical robotic deployment. Rather than focusing on a single improvement, \method advances the system along three functional domains that we find particularly critical for real-world deployment. First, to improve \textbf{\textit{generalization}} across tasks and embodiments, we redesign the data processing pipeline and curate a large-scale pretraining corpus of around 60,000 hours, including 50,000 hours of robot trajectories spanning \robotnum robot configurations and 10,000 hours of egocentric human videos. This design aims to provide broader coverage over both embodiment patterns and interaction scenarios. Second, to support a wider range of practical robots, we extend the model to an \textbf{\textit{expanded action space}} beyond standard dual-arm platforms, enabling control over the head, waist, mobile base, and dexterous hands. This expanded control interface allows the system to address more complex tasks that require coordinated whole-body interaction. Third, to improve temporal reasoning in dynamic environments, as shown in~\cref{fig:teaser}, we introduce \textbf{\textit{predictive dynamics modeling}} as a proxy objective. Concretely, we formulate future prediction using a video representation model to provide semantic priors and a depth estimation model to provide geometric cues, encouraging the VLA model to reason about future scene evolution and action consequences. Our central hypothesis is that practical VLA systems should not only scale in model and data size, but also become better aligned with the demands of real-world robotics: broader embodiment support, richer controllable action spaces, and stronger predictive understanding of dynamic scenes. Under this view, \method is intended as a step from foundation-level capability toward application-oriented usability.

We evaluate \method in a generalist setting on nine tasks from the GM-100 dual-arm manipulation benchmark~\cite{gm100}, as well as on two long-horizon mobile manipulation tasks. The results show that the proposed modifications consistently improve practical capability, validating the importance of jointly enhancing generalization, action-space coverage, and predictive dynamics modeling. Overall, our work highlights a pragmatic direction for advancing VLA foundation models from laboratory success toward real-world applicability.

\begin{figure}[t]
\centering
\includegraphics[width=\linewidth]{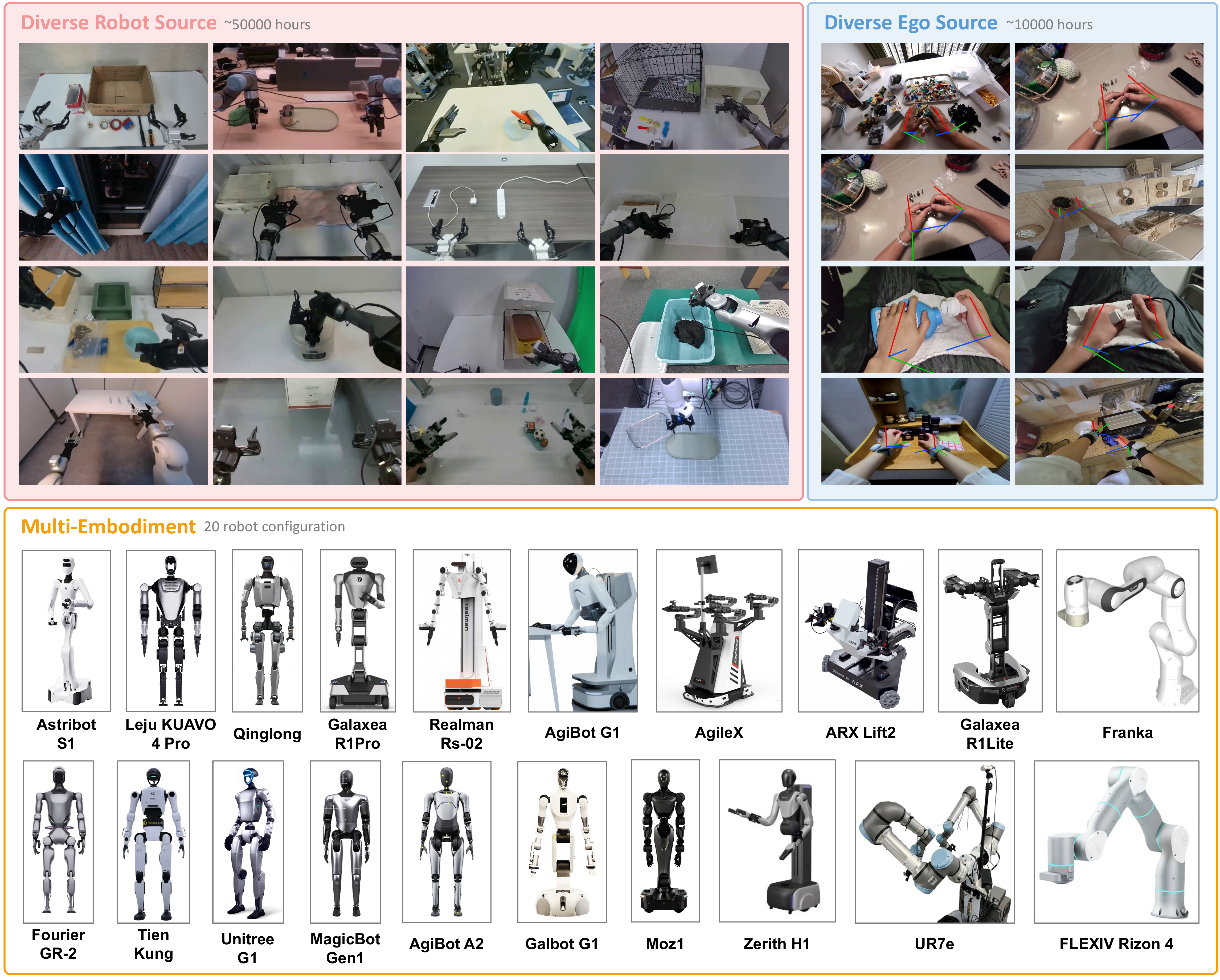}
\caption{\textbf{Visualization of the pre-training dataset used by LingBot-VLA-2.0}. The dataset includes 20 robot embodiments with degrees of freedom in the arms, heads, waists, mobile bases, and dexterous hands.}
\label{fig:data_demo}
\end{figure}

\section{Related Work}\label{sec:related}

\subsection{Generalist Robot Manipulation Policies}
Recent vision-language-action (VLA) models~\cite{pi_0, pi_0d5, gr00tN1, gr00t_n1_6, gemini_robotics, walloss, galaxeaG0, magma, gr3,abotm0,abotm05,hyembodied05x,qwenrobotmanip,qwenVLA,cai2026xiaomi} have established foundation VLA model as a promising paradigm for generalist robot control. Existing efforts have improved VLA systems along several complementary directions, including scaling heterogeneous robot data and unified pretraining recipes for generalization across tasks and embodiments~\cite{qwenVLA,walloss,wallOSS05,starVLAalpha}, leveraging human-centric data such as egocentric videos and hand motion to learn transferable embodied priors~\cite{beingH0,beingH05,beingH07}, introducing embodiment-aware architectures or unified action spaces to better support cross-platform transfer~\cite{holoBrain0}, and incorporating predictive or world-modeling objectives to improve temporal reasoning and future-aware decision making~\cite{lda1b,dexWorldModel}. Other works further enhance VLA models through geometry-aware supervision~\cite{gem}, autoregressive reasoning-action unification~\cite{g0_5}, or deployment-oriented system design for real-world execution~\cite{rldx1,walloss}. Together, these studies demonstrate the rapid progress of foundation VLA models, while also highlighting that practical deployment still requires jointly addressing large-scale generalization, richer embodiment, and action-space support.

\subsection{Mixture-of-Experts Vision-Language-Action Model}
Mixture-of-Experts (MoE) has emerged as a pivotal mechanism for scaling action modeling capacity in VLA systems.
For contact-rich manipulation, ForceVLA~\cite{forcevla} and ForceVLA2~\cite{forcevla2} introduce force-aware MoE modules to fuse sparse but critical force feedback with visual-language representations, while MoDE-VLA~\cite{modevla} further exploits force and tactile signals for dexterous bimanual manipulation. 
For long-horizon tasks, AtomicVLA~\cite{atomicvla} structures experts around atomic skills to mitigate inter-stage interference, while SAMoE-VLA~\cite{samoevla} conditions expert routing on scene-level representations rather than token-level features alone.
Other works~\cite{hex,himoevla, beingH05} introduce MoE structures to address humanoid body-part and motion-phase specialization, and embodiment-level heterogeneity.
In large-scale VLA pretraining, action representations are jointly shaped by multiple entangled factors, including embodiment-specific dynamics, task-dependent control logic, and deployment scenario diversity. Rather than prescribing expert semantics based on any single dimension of variation, we instantiate token-level sparse MoE layers within the action expert, enabling each action token to adaptively select experts based on its intrinsic features. Furthermore, we adopt an auxiliary-loss-free load-balancing mechanism~\cite{liu2024deepseek} that promotes balanced expert utilization via routing correction biases, eliminating the need to inject an explicit load-balancing loss into the primary action learning objective.

\section{Pre-training Dataset}\label{sec:method}

\begin{figure}[t]
\centering
\includegraphics[width=0.7\linewidth]{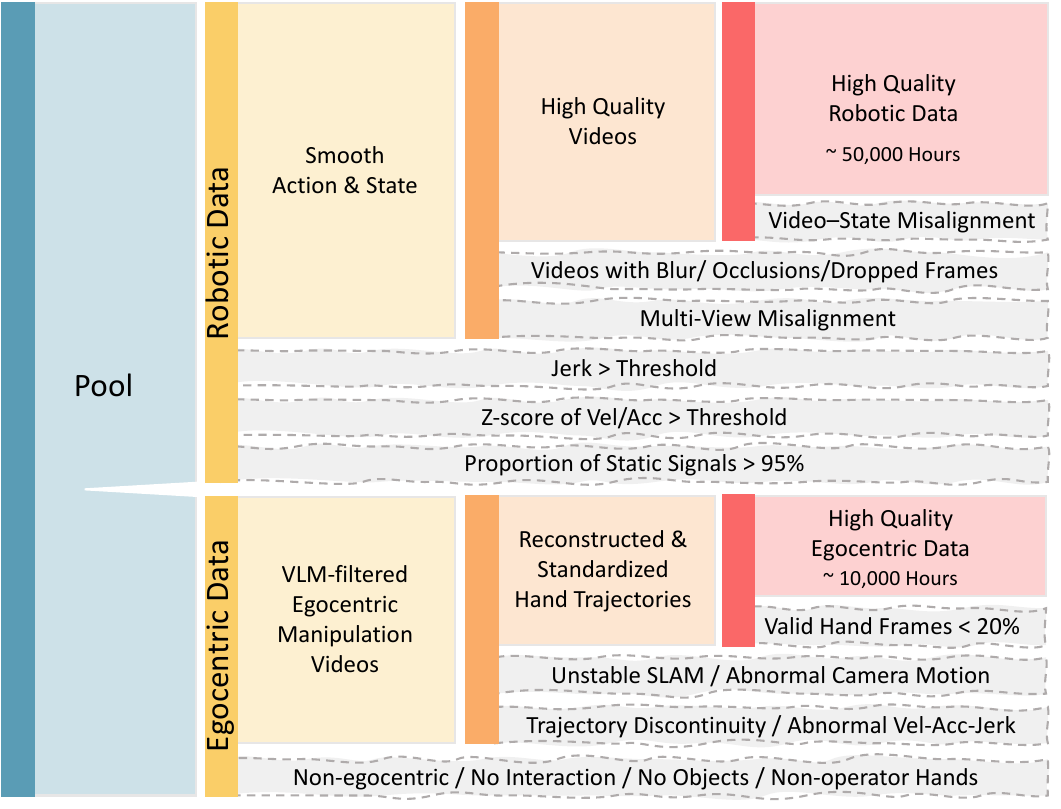}
\caption{Data processing pipeline.}
\label{fig:data_process}
\end{figure}

To improve VLA models for strong generalization and practicality across diverse embodiments and dynamic environments, we redesign the data processing pipeline and curate a large-scale pretraining dataset of approximately 60,000 hours, as illustrated in~\cref{fig:data_demo}. The redesigned data processing pipeline is shown in~\cref{fig:data_process}.

\subsection{Data Curation and Pre-Processing}

\subsubsection{Robotic Data}

We collect approximately 90,000 hours of data from 20 embodiments, spanning single-arm, dual-arm, and mobile robotic platforms equipped with dexterous hands or grippers. Based on the collected data, we employ a redesigned data processing pipeline to filter noisy samples, yielding 50,000 hours of high-quality robotic data.

We first compute the third-order finite difference (jerk) of the action and state signals, along with the Z-scores of their first-order derivatives (velocity) and second-order derivatives (acceleration), to assess trajectory smoothness. An episode is discarded if either the jerk or any derivative Z-score exceeds a predefined threshold. 
These thresholds are set separately for each embodiment.
Moreover, we measure the duration within an episode during which all state and action signals exhibit only small variations or remain unchanged. If this duration accounts for more than 95\% of the entire episode, the episode is also discarded.

To verify consistency between the videos and state signals, we project the robot onto the image plane using the corresponding URDF and replay the recorded states. 
Human annotators are employed to identify discrepancies between the projected robot and the videos to ensure that the state signals and videos are correctly recorded, and samples with such misalignment are removed. Meanwhile, videos with blur, severe occlusions, dropped frames, or misalignment across multiple views are filtered out by human annotators during the annotation process.

\begin{table}[t]
\centering
\caption{Statistical overview of robot manipulation data in LingBot-VLA-2.0.}
\label{tab:linbotvla_robot_data}
\small
\setlength{\tabcolsep}{5pt}
\begin{tabular}{lcccccc}
\toprule
Robot Type & EE type &Hand/Gripper DoF & Arm DoF & Body DoF$^{*}$ & Total DoF & Policy Frequency\\
\midrule
\multicolumn{5}{l}{\textbf{Single-Arm Robots}} \\
Franka & Grppier&1&7&0&8&30 \\
Flexiv Rizon 4 & Grppier  &1&7&0&8&30  \\
\midrule
\multicolumn{5}{l}{\textbf{Dual-Arm Robots}} \\
AgileX & Grppier &2&12&0&14&30\\
ARX Lift2& Grppier &2&12&0&14&30\\
UR7e & Grppier  &2&12&0&14&30  \\
\midrule
\multicolumn{5}{l}{\textbf{Half-Humanoid}} \\
AgiBot G1 & Grppier      &2&14&4&20&30 \\
Galbot G1 & Grppier& 2 & 14 & 6& 22 & 30 \\
Moz1 & Grppier  &2&14&0&16&30  \\ 
Realman Rs-02 & Grppier & 2 & 14 & 1 & 17 & 30 \\
Galaxea R1Pro & Grppier & 2 & 14 & 7 & 23 & 15 \\
Galaxea R1Lite & Grppier & 2 & 12 & 3 & 17 & 15 \\
Astribot S1 & Grppier & 2 & 14 & 9 & 25 & 30 \\
Zerith H1 & Grppier & 2 & 14 & 7 & 23 & 30 \\
\midrule
\multicolumn{5}{l}{\textbf{Humanoid}} \\
Leju KUAVO 4 Pro & Grppier/Hand&2/12&14&5&21/31&30 \\
Tienkung & Grppier& 2& 14 & 0 & 16 & 30\\ 
QingLong & Grppier& 2& 14 & 0 & 16 & 30\\ 
MagicBot Gen1&Grppier& 2& 14 & 4 & 20 & 30 \\
Unitree G1 & Hand& 12 & 14 & 0 & 26 & 30  \\
Fourier GR-2 &Hand & 12 & 14 & 6 & 32 &30\\
AgiBot A2 &Hand & 12 & 14 & 2 & 28 & 30 \\
\midrule
\multicolumn{5}{l}{\textbf{Humanoid}} \\
Ego & - &- &14 &- &14  &30\textasciitilde 60 \\
\midrule
\multicolumn{6}{l}{\textbf{Total: 20 embodiments}} & \textbf{60000h} \\
\bottomrule
\end{tabular}
\begin{tablenotes}
\footnotesize
\item[] $*$ The total number of body DoF used for model training.
\end{tablenotes}
\end{table}

\subsubsection{Egocentric Data}

We construct an egocentric human video pool of approximately 20,000 hours and retain around 10,000 hours of high-quality training data after filtering, reconstruction, standardization, and quality control. 
We first apply a unified video-level VLM pre-filter to all candidate videos, regardless of source. This step removes videos that do not satisfy the egocentric manipulation assumption, such as third-person observation videos, scene-walking videos, videos without clear hand-object interaction, or videos without manipulable objects. We also filter out videos where non-operator hands appear prominently, since such clips can break the association between the camera wearer's observation and the corresponding hand motion. Applying this VLM filtering stage before downstream processing reduces unnecessary annotation and reconstruction cost, especially for action-free videos that would otherwise require SLAM and hand pose estimation.

After VLM pre-filtering, we process the remaining data according to whether action or hand trajectory labels are available. For data with existing action or hand trajectory labels, including action-labeled open-source datasets and in-house data, we perform metadata organization, timestamp alignment, coordinate transformation, and trajectory completeness checking, converting all samples into a standardized hand trajectory format. For action-free egocentric human videos, we run egocentric SLAM to estimate camera intrinsics and per-frame camera extrinsics, and then apply hand pose estimation to recover MANO parameters in the camera coordinate frame. By combining the estimated hand poses with camera poses, we lift the hand motion into the world coordinate frame and obtain temporally continuous hand trajectories.

We then apply trajectory-level quality control to the reconstructed and standardized data. We filter out videos with insufficient valid hand pose coverage, using a 20\% valid-frame ratio as the minimum threshold. We also remove clips with unstable SLAM trajectories, identified by abnormal second-order changes in the estimated camera motion, such as sudden translational or rotational acceleration. In addition, we reject hand trajectories with sudden discontinuities, abnormal displacement, velocity, acceleration, or jerk, as well as samples that violate human physiological constraints, such as unreasonable hand positions, inter-hand distances, motion ranges, or poses.

Finally, all valid samples are stored as hand trajectories in the world coordinate frame. During training, when a frame is sampled as the current observation, we transform the future hand trajectory from the world coordinate frame into the current camera coordinate frame using the camera extrinsic of that frame:
\begin{equation}
    \mathbf{p}_{\tau}^{C_t} = \mathbf{T}_{C_t \leftarrow W} \mathbf{p}_{\tau}^{W}.
\end{equation}
Here, $\mathbf{p}_{\tau}^{W}$ denotes the hand trajectory in the world coordinate frame, and $\mathbf{T}_{C_t \leftarrow W}$ denotes the transformation from the world coordinate frame to the camera coordinate frame of the sampled frame $t$. With this design, the world coordinate frame serves as the unified trajectory storage space, while the current camera coordinate frame serves as the training-time action representation. This unifies trajectory formats across open-source and in-house egocentric human videos, and decouples hand motion from egocentric camera motion during training.

\subsubsection{Unified Action Representation} 

\begin{figure}[t]
\centering
\includegraphics[width=\linewidth]{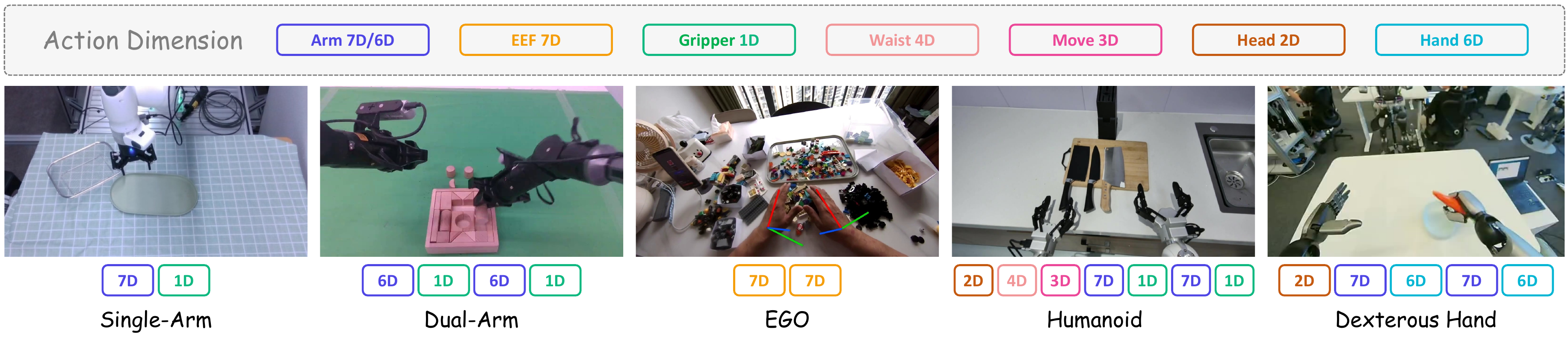}
\caption{\textbf{Unified action representation}. We map heterogeneous embodiment controls into compact action vectors composed of shared body-part components.}
\label{fig:data_dimension}
\end{figure}

To jointly learn state and action representations from egocentric data and robotic data collected across multiple embodiments, we use a 55-dimensional canonical vector representation for both states and actions. This representation consists of 14 dimensions for arm joint position, 14 dimensions for end-effector pose, 2 dimensions for gripper position, 12 dimensions for hand joint position, 4 dimensions for waist position, 2 dimensions for head position, and 3 dimensions for mobility signal, as shown in~\cref{fig:data_dimension}. The remaining 4 dimensions are reserved.
The arm joint position and end-effector pose fields are defined to accommodate the maximum dimensionality of a dual-arm embodiment. Specifically, the end-effector pose of each arm is represented by XYZ coordinates and a rotation quaternion, resulting in 7 dimensions per arm. For single-arm data, only 6 or 7 dimensions are used for arm joint positions and 7 dimensions for the end-effector pose, while the remaining arm-related dimensions are padded.
Similarly, for robot embodiments that do not include specific body parts or have lower-dimensional signals, the corresponding dimensions are also padded.
The dimensional configurations for each embodiment are shown in~\cref{tab:linbotvla_robot_data}.

\subsection{Data Annotation}
\label{sec:data_annotation}

VLA pretraining benefits from manipulation videos paired with temporally aligned language supervision at both the task and subtask levels. We generate these annotations with a fully automated pipeline built on a vision--language model, and apply this pipeline across the pretraining dataset.

\begin{figure}[t]
\centering
\includegraphics[width=0.8\linewidth]{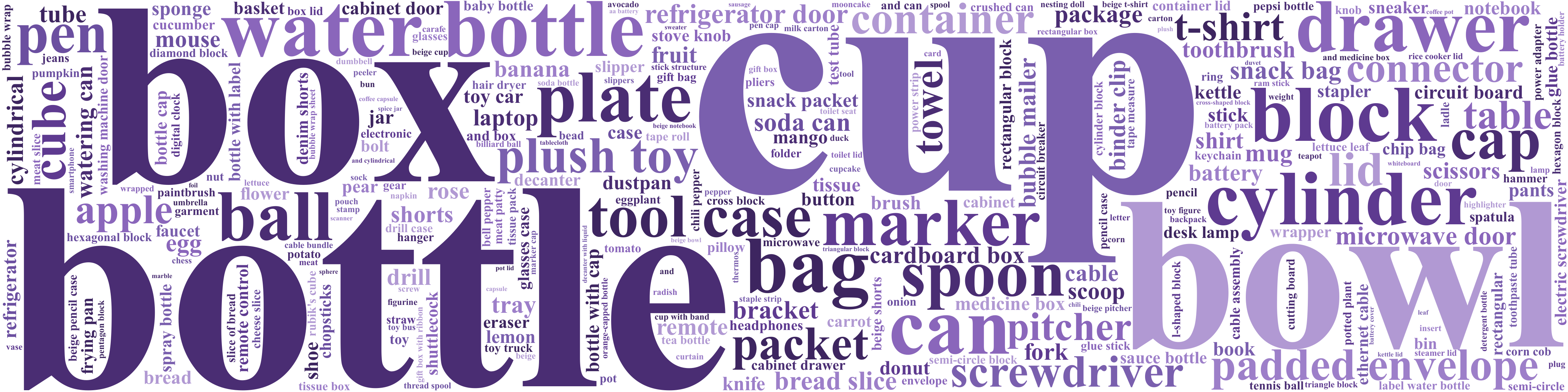}
\caption{Word cloud of the manipulated objects in the subtask annotations, sized by frequency.}
\label{fig:object_cloud}
\end{figure}

Specifically, we employ Qwen3.6-27B~\cite{qwen3.6-27b} to segment each manipulation video into a sequence of temporally contiguous subtasks and to generate the corresponding language instructions. For robotic platforms with multiple cameras, the overhead and wrist views are processed jointly to disambiguate gripper--object interactions. Each subtask is assigned an atomic action from a closed vocabulary of 18 categories (\cref{tab:subtask_vocab}), together with the primary object of interaction and a concise instruction. In addition, a single video-level instruction summarizes the overall task. The annotated objects form a diverse, open vocabulary, as visualized in~\cref{fig:object_cloud}.

\begin{figure}[t]
\centering
\includegraphics[width=\linewidth]{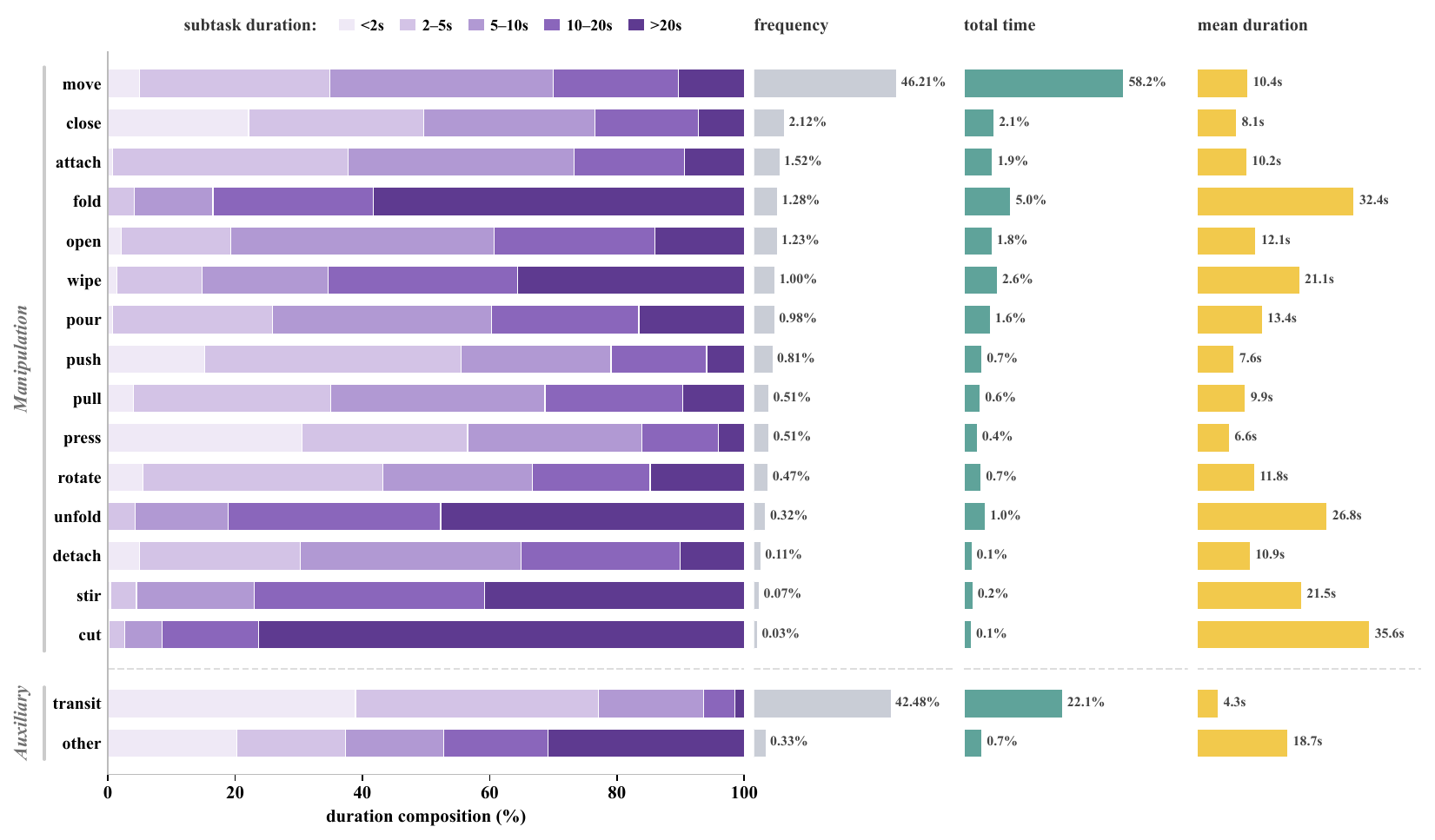}
\caption{Per-action statistics of the subtask annotations: duration composition, \emph{frequency} (fraction of all subtasks), \emph{total time} (fraction of the total annotated time), and \emph{mean duration}. Actions are grouped as in~\cref{tab:subtask_vocab}; \emph{idle} is filtered out of the training data and omitted here.}
\label{fig:subtask_stats}
\end{figure}

To keep the segmentation granularity consistent, the model groups the grasp, carry, and release phases of a single interaction into one subtask, and introduces a temporal boundary only when the manipulated object changes, the action type changes, or a sustained pause marks a transition to a new sub-goal. As summarized in~\cref{fig:subtask_stats}, \emph{move} and \emph{transit} dominate in frequency, whereas fine-grained manipulations such as \emph{cut}, \emph{fold}, and \emph{stir} are rare but have considerably longer mean durations.

\begin{table}[t]
\centering
\small
\setlength{\tabcolsep}{5pt}
\renewcommand{\arraystretch}{1.08}
\caption{
Closed vocabulary for subtask annotation.
The vocabulary consists of 15 primitive manipulation actions and three auxiliary labels:
\texttt{transit} for empty-hand motion, \texttt{idle} for stationary arms, and
\texttt{other} for out-of-vocabulary actions.
}
\label{tab:subtask_vocab}
\begin{tabularx}{\columnwidth}{@{}>{\ttfamily}l X @{\quad} >{\ttfamily}l X@{}}
\toprule
\normalfont\textbf{Action} & \textbf{Description}
& \normalfont\textbf{Action} & \textbf{Description} \\
\midrule

\multicolumn{4}{@{}l}{\textbf{Primitive manipulation actions}} \\
\addlinespace[0.15em]
move   & Relocate an object
& fold   & Shrink flexible material \\
pour   & Tilt a container to dispense contents
& unfold & Expand flexible material \\
push   & Slide an unheld object
& wipe   & Move a tool across a surface \\
pull   & Draw an object closer
& stir   & Move a tool circularly in a container \\
rotate & Turn an object in place
& cut    & Sever a target with a blade \\
open   & Open a hinged or articulated object
& press  & Press a button, switch, or surface \\
close  & Close a hinged or articulated object
& attach & Insert or connect an object to a fitting \\
detach & Remove or disconnect an object from a fitting
&        & \\

\addlinespace[0.35em]
\midrule
\multicolumn{4}{@{}l}{\textbf{Auxiliary labels}} \\
\addlinespace[0.15em]
transit & Move the gripper without object interaction
& idle   & Keep the arms stationary \\
other   & Action outside the vocabulary
&        & \\

\bottomrule
\end{tabularx}
\end{table}

\section{Method}

\subsection{MoE-based VLA model}
Vision-Language-Action (VLA) models pretrained on real-world cross-embodiment robot data must learn from heterogeneous trajectories characterized by misaligned action spaces, varying embodiment dynamics, and task distributions. Although the Mixture-of-Experts (MoE) mechanism provides a natural way to scale model capacity, the underlying reasons for its effectiveness in VLA pretraining remain poorly understood. In this report, we introduce a token-level loss-free MoE architecture for large-scale multi-embodiment VLA pretraining. By decoupling expert load balancing from the primary action-learning objective, and employing sigmoid-based routing confidence to allow each token to independently activate multiple experts, our design enables sparse experts to capture general representations and control logic shared across embodiments.

\noindent\textbf{Sparse MoE Architecture.}
\label{sec:moe_based_vla_model}
To facilitate convergence during cross-embodiment pretraining, we instantiate sparse MoE layers inside the action expert, which replace the feed-forward network (FFN) in the action expert. In our default configuration, all action-expert transformer blocks are equipped with MoE FFNs. Furthermore, we adopt fine-grained expert segmentation and shared expert isolation, in which a lightweight shared expert preserves universal priors, while a set of routed experts provides specialized modeling capacity.

Given the modulated FFN input $u_{\ell,t} \in \mathbb{R}^{d}$ of token $t$ at layer $\ell$, the MoE layer computes
\begin{equation}
m_{\ell}(u_{\ell,t}) =
E_{\ell}^{(s)}(u_{\ell,t})
+
\lambda
\sum_{j \in \mathcal{R}(u_{\ell,t})}
g_{\ell,j}(u_{\ell,t}) E_{\ell,j}^{(r)}(u_{\ell,t}),
\end{equation}
where $E_{\ell}^{(s)}(\cdot)$ denotes the shared expert, $E_{\ell,j}^{(r)}(\cdot)$ denotes the $j$-th routed expert, $\mathcal{R}(u_{\ell,t})$ is the selected top-$K$ routed expert set, and $\lambda$ is a routed-output scaling factor. Each shared or routed expert is implemented as a SwiGLU MLP:
\begin{equation}
E(u) = W_{\mathrm{down}}
\left(
\mathrm{SiLU}(W_{\mathrm{gate}}u) \odot W_{\mathrm{up}}u
\right).
\end{equation}
Compared with the dense FFN counterpart, both the shared and routed experts employ a smaller intermediate width, encouraging shared experts to capture general principles and routed experts to provide stronger specialization. In practice, we use one shared expert and $N_r$ routed experts per MoE layer; only $K$ routed experts are activated for each token.
For token-choice routing, we compute router logits with a linear router in FP32:
\begin{equation}
z_{\ell,j}(u_{\ell,t}) = u_{\ell,t}^{\top} e_{\ell,j},
\end{equation}
where $e_{\ell,j}$ is the learnable router embedding of the $j$-th routed expert. To avoid the strong competition among experts induced by softmax normalization, we apply a sigmoid-activated function to the router logits following DeepSeek-V3~\cite{liu2024deepseek}, the token-to-expert affinity is defined as
\begin{equation}
s_{\ell,j}(u_{\ell,t}) = \operatorname{Sigmoid}(z_{\ell,j}(u_{\ell,t})).
\end{equation}

To preserve the primary objective of action control learning in VLA models, we adopt an auxiliary-loss-free strategy inspired by DeepSeek-V3~\cite{liu2024deepseek} to promote load balancing within the MoE architecture.
Under auxiliary-loss-free balancing, each expert maintains a routing correction bias $b_{\ell,j}$. The actual mixture weights are still computed from the original unbiased affinities:
\begin{equation}
g_{\ell,j}(u_{\ell,t}) =
\frac{s_{\ell,j}(u_{\ell,t})}
{\sum_{k \in \mathcal{R}(u_{\ell,t})} s_{\ell,k}(u_{\ell,t})},
\quad j \in \mathcal{R}(u_{\ell,t}),
\end{equation}
while the selected routed expert set is determined by biased affinities:
\begin{equation}
\mathcal{R}(u_{\ell,t}) =
\mathrm{TopK}_{j}
\left(
s_{\ell,j}(u_{\ell,t}) + b_{\ell,j}, K
\right).
\end{equation}
This bias-based balancing mechanism decouples expert assignment from expert routing. Specifically, the correction bias is employed to promote load balancing, while the routing confidence remains weighted by the model's original, unbiased affinity scores.
During training, we accumulate the number of tokens assigned to each expert across micro-batches and distributed ranks. At each load-balancing update, the correction bias is adjusted according to the sign of the deviation from the mean expert load:
\begin{equation}
b_{\ell,j} \leftarrow
b_{\ell,j}
-
\gamma \cdot
\mathrm{sign}
\left(
n_{\ell,j} - \frac{1}{N_r}\sum_{k=1}^{N_r} n_{\ell,k}
\right),
\end{equation}
where $n_{\ell,j}$ is the accumulated load of expert $j$ in layer $\ell$, and $\gamma$ is the bias update speed. Optionally, the bias vector is centered after each update to prevent cumulative drift. The MoE output is then injected back through the original FFN residual branch of the action expert transformer.

\begin{figure}[H]
    \centering
    \includegraphics[width=0.96\linewidth]{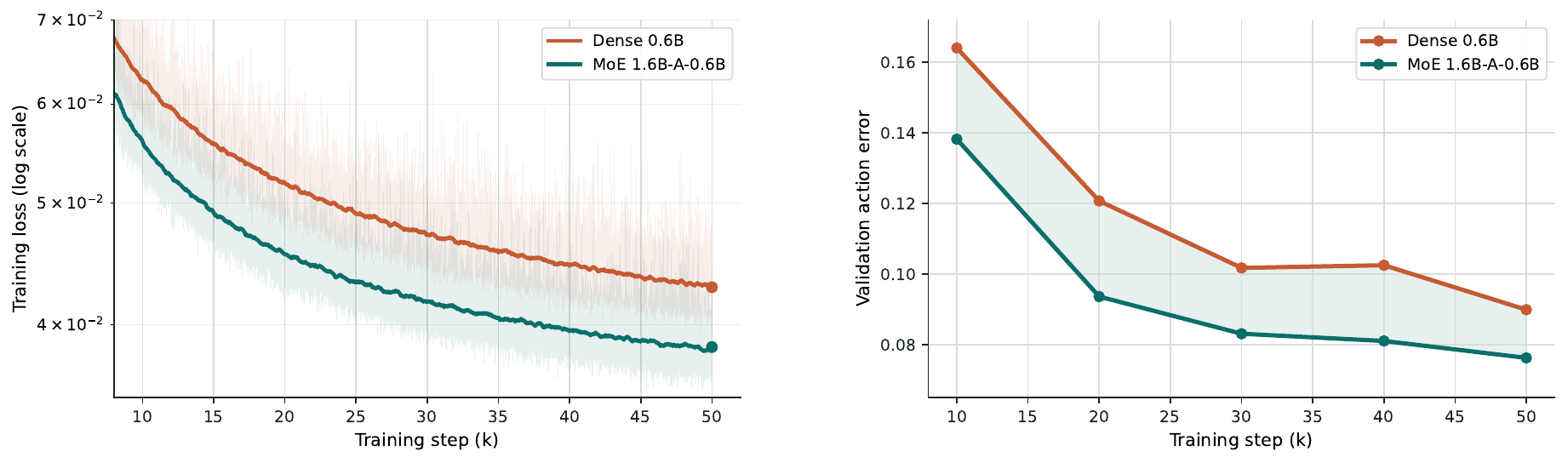}
    \caption{Comparable active-parameter scaling comparison between Dense model and MoE model using training loss on pre-training data and validation action error on GM-100 tasks.}
    \label{fig:vla_loss_dense_vs_moe}
\end{figure}

\noindent\textbf{Scaling Experiments.}
To validate the efficacy of the sparse architecture under an identical compute budget, we conduct a comprehensive comparison between MoE and Dense models with strictly matched active parameter counts. Experimental results shown in \cref{fig:vla_loss_dense_vs_moe} indicate that MoE consistently achieves lower training loss and validation error than its dense counterpart. 
This advantage is observed across both optimization and generalization metrics, indicating that the performance gain of MoE does not merely arise from increased total parameter count, but from a more effective allocation of model capacity through sparse activation. These results demonstrate that, under a fixed compute budget, the sparse MoE architecture achieves more efficient scaling with reduced training loss and validation error, establishing it as a more effective scaling strategy for VLA pre-training.

\subsection{Spatiotemporal-Aware VLA via Dual-Query Distillation}
To promote both geometric awareness and causal temporal understanding in the \method, we adopt a dual-query distillation framework inspired by recent works~\cite{wang2025vision}, where LingBot-Depth and our robotics-aware DINO-Video serve as complementary teachers.
Specifically, we append two learnable queries, $[\mathbf{Q}_t, \mathbf{Q}_{t+T}]$, to the visual and textual tokens, where $\mathbf{Q}_t$ targets the current observation and $\mathbf{Q}_{t+T}$ targets the future observation at horizon $T$ (\textit{i.e.}, the action chunk size). These contextualized query states are then distilled from two complementary teachers: a depth teacher, LingBot-Depth~\cite{lingbotdepth}, which provides explicit geometric supervision, and a causal video teacher, \dinomodel, which provides temporally grounded visual representations.

\noindent\textbf{LingBot-Depth for geometric supervision.}
To inject explicit spatial and geometric priors into \method, we align the dual-query states with depth tokens extracted from LingBot-Depth. Specifically, the current and future queries, $[\mathbf{Q}_t, \mathbf{Q}_{t+T}]$, are trained to predict the corresponding depth representations $[\mathbf{D}_t, \mathbf{D}_{t+T}]$ from the current and future manipulation frames, respectively. The depth distillation objective is
\begin{align}
\mathcal{L}_{depth}
=
\mathbb{E}\left[
\left\|\mathrm{Proj}_{depth}(\mathbf{Q}_t)-\mathbf{D}_t\right\|_1
+
\left\|\mathrm{Proj}_{depth}(\mathbf{Q}_{t+T})-\mathbf{D}_{t+T}\right\|_1
\right],
\label{eq:depth_distill}
\end{align}
where $\mathrm{Proj}_{depth}(\cdot)$ denotes a projection module with cross-attention for dimensional alignment. 
Within the causal VLM architecture, $\mathbf{Q}_t$ captures the immediate scene geometry, while $\mathbf{Q}_{t+T}$ learns to anticipate future geometric configurations relevant to upcoming manipulation.

\noindent\textbf{Causal \dinomodel for temporal supervision.}
While depth supervision provides geometric structure, it does not capture the causal temporal dynamics required for robotic control. Motivated by the success of latent visual representations across a broad range of action and robotics applications~\cite{larybench, beingH07, lda1b, dexWorldModel, lawam, wog}, we introduce \dinomodel, a robotics-aware video representation model built on top of the DINOv3~\cite{dinov3} image backbone. Unlike image-level features extracted independently for each frame, \dinomodel produces motion-aware visual representations with causal temporal attention, such that the feature at each time step depends only on the current and past observations.
Given the contextualized query states $[\mathbf{Q}_t, \mathbf{Q}_{t+T}]$, the current query is trained to predict the \dinomodel feature of the current frame, while the future query is trained to predict the feature of the future frame at horizon $T$. Denoting the teacher targets by $[\mathbf{Z}_t, \mathbf{Z}_{t+T}]$, extracted from a single causal forward pass of \dinomodel over the corresponding observation clip, the video distillation objective is
\begin{align}
\mathcal{L}_{video}
=
\mathbb{E}\left[
\left\|\mathrm{Proj}_{video}(\mathbf{Q}_t)-\mathbf{Z}_t\right\|_F^2
+
\left\|\mathrm{Proj}_{video}(\mathbf{Q}_{t+T})-\mathbf{Z}_{t+T}\right\|_F^2
\right],
\label{eq:dino_video}
\end{align}
where $\mathrm{Proj}_{video}(\cdot)$ maps the query states into the patch-level feature space of \dinomodel. This objective encourages \method to recover both the current motion-aware representation and its future counterpart, complementing the geometric supervision from depth distillation.

\noindent\textbf{Details of robotics-aware \dinomodel.}
We initialize \dinomodel from DINOv3 and extend it with block-wise causal temporal attention and 3D rotary positional embeddings (3D-RoPE)~\cite{videorope}, enabling causal video modeling while preserving the spatial semantics and geometry priors inherited from DINOv3. 
We train \dinomodel on 5M video clips spanning Internet, egocentric, and robotic data, using video-adapted DINO and iBOT self-distillation objectives~\cite{omnistream}. 
For each sample, we uniformly sample 16 frames and assign an absolute temporal encoding based on the effective frame rate to distinguish clips with different real-time spans~\cite{cosmos3}. 
On LARYBench~\cite{larybench}~(see \cref{tab:larybench}), \dinomodel achieves the best performance on three of four benchmarks, supporting its effectiveness as a robotics-aware temporal teacher.

\begin{table*}[h]
\centering
\caption{Results of \dinomodel on LARYBench Classification and Regression Evaluation.}
\label{tab:larybench}
\begin{tabular}{llcccc}
\toprule
\multirow{2}{*}{Model} & \multirow{2}{*}{Params(M)} & \multicolumn{2}{c}{Classification}  & \multicolumn{2}{c}{Regression} \\
 & & \textit{Composite Human}$\uparrow$ & \textit{Composite Robot}$\uparrow$ & \textit{RoboCOIN}$\downarrow$ & \textit{AgiBotWorld-Beta}$\downarrow$ \\
\midrule
V-JEPA 2~\cite{vjepa2} & 303.89 &\textbf{ 80.35} & 70.43 & 0.32 & 0.33 \\
DINOv3~\cite{dinov3} & 303.13 & 76.19 & 69.06 & 0.22 & 0.24\\

\midrule
\dinomodel & 303.13 & {80.21} & \textbf{71.97} & \textbf{0.20} & \textbf{0.19}\\
\bottomrule

\end{tabular}
\end{table*}

\begin{table}[ht!]
\centering
\scriptsize
\setlength{\tabcolsep}{3pt}
\renewcommand{\arraystretch}{1.1}
\caption{Scoring criteria for the part of GM-100 benchmark tasks.}
\label{tab:bm_scoring}
\begin{tabular}{p{0.12\columnwidth} p{0.22\columnwidth} p{0.56\columnwidth}}
\toprule
\textbf{Task ID} & \textbf{Task} & \textbf{Scoring Criteria} \\
\midrule

BM-19 & Block Sorting &
\makecell[l]{
1. Pick up the largest block on the table (16). \\
2. Place the largest block on the far left (12). \\
3. Pick up the second-largest block on the table (12). \\
4. Place the second-largest block to the right of the largest block (12). \\
5. Pick up the third-largest block on the table (12). \\
6. Place the third-largest block to the right of the second-largest block (12). \\
7. Pick up the smallest block on the table (12). \\
8. Place the smallest block on the far right (12).
} \\
\midrule

BM-20 & Retrieve Keychain &
\makecell[l]{
1. Pull open the drawer (25). \\
2. Grasp the keychain (25). \\
3. Move the keychain to the front of the drawer (25). \\
4. Put down the keychain (25).
} \\
\midrule

BM-25 & Scoop Rice &
\makecell[l]{
1. Pick up the spoon with the right hand (20). \\
2. Scoop rice from the bowl with the spoon (30). \\
3. Pour the rice into the cup (30). \\
4. Place the spoon back onto the plate (20).
} \\
\midrule

BM-39 & Replace Paper Roll &
\makecell[l]{
1. Remove the empty paper-roll core (22). \\
2. Place the core on the tray (22). \\
3. Pick up the new paper roll (22). \\
4. Install the new paper roll onto the holder (34).
} \\
\midrule

BM-45 & Sort Snacks &
\makecell[l]{
1. Pick up the first type of snack (18). \\
2. Place it into the corresponding container (16). \\
3. Pick up the second type of snack (17). \\
4. Place it into the corresponding container (16). \\
5. Pick up the third type of snack (17). \\
6. Place it into the corresponding container (16).
} \\
\midrule

BM-47 & Pack Eggs &
\makecell[l]{
1. Pick up the first egg (16). \\
2. Place the egg into a tray slot (12). \\
3. Pick up the second egg (12). \\
4. Place the egg into a tray slot (12). \\
5. Pick up the third egg (12). \\
6. Place the egg into a tray slot (12). \\
7. Pick up the fourth egg (12). \\
8. Place the egg into a tray slot (12).
} \\
\midrule

BM-69 & Pick Out Toy Bones &
\makecell[l]{
1. Pick up the first toy bone (25). \\
2. Move the bone out of the plate (25). \\
3. Pick up the second toy bone (25). \\
4. Move the bone out of the plate (25).
} \\
\midrule

BM-75 & Push Ball into Box &
\makecell[l]{
1. Push the ball toward the box (30). \\
2. Push the ball into the box (70).
} \\
\midrule

BM-82 & Squeeze Ketchup &
\makecell[l]{
1. Pick up the ketchup bottle with the left hand and hold it above the plate (25). \\
2. Tilt the bottle so that the nozzle points toward the plate (25). \\
3. Squeeze the bottle with both hands to dispense ketchup (25). \\
4. Put down the ketchup bottle (25).
} \\
\midrule

BM-97 & Take Bowl out of Microwave &
\makecell[l]{
1. Open the microwave door (25). \\
2. Grasp the bowl inside the microwave (25). \\
3. Move the bowl onto the table and place it down (25). \\
4. Close the microwave door (25).
} \\
\midrule

BM-105 & Tool Packing &
\makecell[l]{
1. Pick up the tape and place it into the toolbox (20). \\
2. Pick up the measuring tape and place it into the toolbox (20). \\
3. Pick up the screwdriver and place it into the toolbox (20). \\
4. Pick up the cutter and place it into the toolbox (20). \\
5. Close the toolbox (20).
} \\
\midrule

BM-107 & Barcode Scan &
\makecell[l]{
1. Lift the medicine box and hold it up with the left hand (20). \\
2. Pick up the barcode scanner with the right hand (20). \\
3. Scan the barcode on the medicine box with the scanner (20). \\
4. Put down the barcode scanner (20). \\
5. Put the medicine box back down (20).
} \\

\bottomrule
\end{tabular}
\end{table}

\section{Experiments}\label{sec:exp}

\subsection{Experimental Settings}

\noindent\textbf{Scoring criteria.} 
\Cref{tab:bm_scoring} summarizes the scoring criteria for a subset of GM-100 benchmark tasks. Each task is decomposed into a sequence of fine-grained manipulation steps, and a partial score is assigned to each step according to task completion. These criteria cover diverse bimanual capabilities, including sorting, retrieval, scooping, replacement, packing, pushing, squeezing, and coordinated object handling. Such stepwise annotations enable a more fine-grained evaluation than binary success alone, capturing partial progress on long-horizon manipulation tasks.

\subsection{Bimanual Manipulation Experiment Results}

\begin{table*}[h]
\centering
\small
\setlength{\tabcolsep}{4.2pt}
\renewcommand{\arraystretch}{1.08}
\caption{
Performance on the bimanual GM-100 benchmark under the generalist setting.
We report progress score (Prog., \%) and success rate (Succ., \%) for each task.
The shaded row in each platform block reports the average over the nine tasks.
}
\label{tab:bimanual_combined}
\begin{tabular}{@{}lrrrrrrrr@{}}
\toprule
\multirow{2}{*}{Task} &
\multicolumn{2}{c}{GR00T N1.7} &
\multicolumn{2}{c}{$\pi_{0.5}$} &
\multicolumn{2}{c}{LingBot-VLA-1.0} &
\multicolumn{2}{c}{LingBot-VLA-2.0} \\
\cmidrule(lr){2-3}
\cmidrule(lr){4-5}
\cmidrule(lr){6-7}
\cmidrule(l){8-9}
& Prog. & Succ. & Prog. & Succ. & Prog. & Succ. & Prog. & Succ. \\
\midrule

\multicolumn{9}{@{}l}{\textbf{Agilex Cobot Magic}} \\
\rowcolor{gray!8}
\textbf{Overall average}
& 36.3 & 17.8 & 59.1 & 32.2 & 58.2 & 30.0 & \textbf{66.2} & \textbf{34.4} \\

Block sorting              
& 40.0 & 10.0 & 90.4 & 60.0 & 59.2 & 10.0 & 56.8 & 0.0 \\
Retrieve keychain          
& 12.5 & 10.0 & 20.0 & 20.0 & 67.5 & 60.0 & 100.0 & 100.0 \\
Replace paper roll         
& 52.8 & 0.0 & 62.8 & 10.0 & 59.6 & 20.0 & 55.2 & 20.0 \\
Sort snacks                
& 91.9 & 70.0 & 82.4 & 30.0 & 74.4 & 10.0 & 66.2 & 10.0 \\
Pack eggs                  
& 14.4 & 0.0 & 72.4 & 20.0 & 42.4 & 10.0 & 44.4 & 0.0 \\
Pick out toy bone          
& 70.0 & 60.0 & 100.0 & 100.0 & 77.5 & 70.0 & 95.0 & 90.0 \\
Push ball into box         
& 18.0 & 0.0 & 38.0 & 20.0 & 6.0 & 0.0 & 41.0 & 20.0 \\
Take bowl out of microwave 
& 15.0 & 10.0 & 0.0 & 0.0 & 77.5 & 70.0 & 77.5 & 70.0 \\
Tool packing               
& 12.0 & 0.0 & 66.0 & 30.0 & 60.0 & 20.0 & 60.0 & 0.0 \\

\addlinespace[0.35em]
\midrule

\multicolumn{9}{@{}l}{\textbf{Galaxea R1 Pro}} \\
\rowcolor{gray!8}
\textbf{Overall average}
& 16.4 & 5.6 & 27.4 & 8.9 & 32.7 & \textbf{15.6} & \textbf{34.6} & \textbf{15.6} \\

Block sorting              
& 8.4 & 0.0 & 12.4 & 0.0 & 28.0 & 0.0 & 33.2 & 0.0 \\
Retrieve keychain          
& 0.0 & 0.0 & 0.0 & 0.0 & 0.0 & 0.0 & 0.0 & 0.0 \\
Replace paper roll         
& 6.6 & 0.0 & 72.0 & 50.0 & 44.0 & 0.0 & 57.6 & 40.0 \\
Sort snacks                
& 31.3 & 0.0 & 28.1 & 0.0 & 10.6 & 0.0 & 26.1 & 0.0 \\
Pack eggs                  
& 1.6 & 0.0 & 6.7 & 0.0 & 19.2 & 0.0 & 4.4 & 0.0 \\
Pick out toy bone          
& 77.5 & 50.0 & 72.5 & 30.0 & 62.5 & 40.0 & 87.5 & 70.0 \\
Push ball into box         
& 9.0 & 0.0 & 21.0 & 0.0 & 3.0 & 0.0 & 18.0 & 0.0 \\
Take bowl out of microwave 
& 7.5 & 0.0 & 7.5 & 0.0 & 97.5 & 100.0 & 65.0 & 30.0 \\
Tool packing               
& 6.0 & 0.0 & 26.0 & 0.0 & 30.0 & 0.0 & 20.0 & 0.0 \\

\bottomrule
\end{tabular}
\end{table*}

\noindent\textbf{Experiment settings.}
We evaluate bimanual manipulation on nine tasks from the GM-100 benchmark under a generalist mixed-training setting. Unlike task-specific training, a single policy is jointly trained on all tasks for each embodiment, requiring the model to share manipulation primitives across diverse bimanual skills, including sorting, retrieval, packing, pushing, and articulated-object interaction. We report the progress score and success rate for each task.

\noindent\textbf{Results analysis.}
As shown in~\cref{tab:bimanual_combined}, LingBot-VLA-2.0 achieves the best overall performance under the generalist setting. On Agilex Cobot Magic, it reaches 66.2 / 34.4 in progress score / success rate, surpassing LingBot-VLA-1.0 by 8.0 / 4.4 points and $\pi_{0.5}$ by 7.1 / 2.2 points. On Galaxea R1 Pro, it achieves 34.6 / 15.6, outperforming $\pi_{0.5}$ by 7.2 / 6.7 points. These results demonstrate that LingBot-VLA-2.0 benefits from mixed-task training and acquires more generalizable bimanual manipulation behaviors.

The improvements are most pronounced on tasks that demand accurate object grounding and goal-directed action execution. For instance, on Agilex \textit{Retrieve keychain}, LingBot-VLA-2.0 improves over LingBot-VLA-1.0 from 67.5 / 60.0 to 100.0 / 100.0, and on Agilex \textit{Pick out toy bone}, from 77.5 / 70.0 to 95.0 / 90.0. A similar gain is observed on the Galaxea \textit{Pick out toy bone} task, where the performance increases from 62.5 / 40.0 to 87.5 / 70.0. These improvements align well with the design of LingBot-VLA-2.0, which adopts a stronger VLM backbone with enhanced grounding capability and conditions action prediction on future information, together enabling more object-centric and goal-consistent manipulation.

Nevertheless, the gains are not uniform across tasks. Several tasks still exhibit a substantial gap between progress score and success rate, indicating that the model often makes partial progress but fails at the final precise placement, release, or completion step. Moreover, the performance disparity between Agilex Cobot Magic and Galaxea R1 Pro suggests that embodiment-specific factors, such as kinematics, camera viewpoints, and action-space alignment, remain challenging. Overall, LingBot-VLA-2.0 delivers consistent improvements in generalist bimanual manipulation, particularly on tasks requiring strong visual grounding and future-aware action planning.

\subsection{Long-Horizon Mobile Manipulation Experiment Results}

\begin{figure}[t]
\centering
\includegraphics[width=\linewidth]{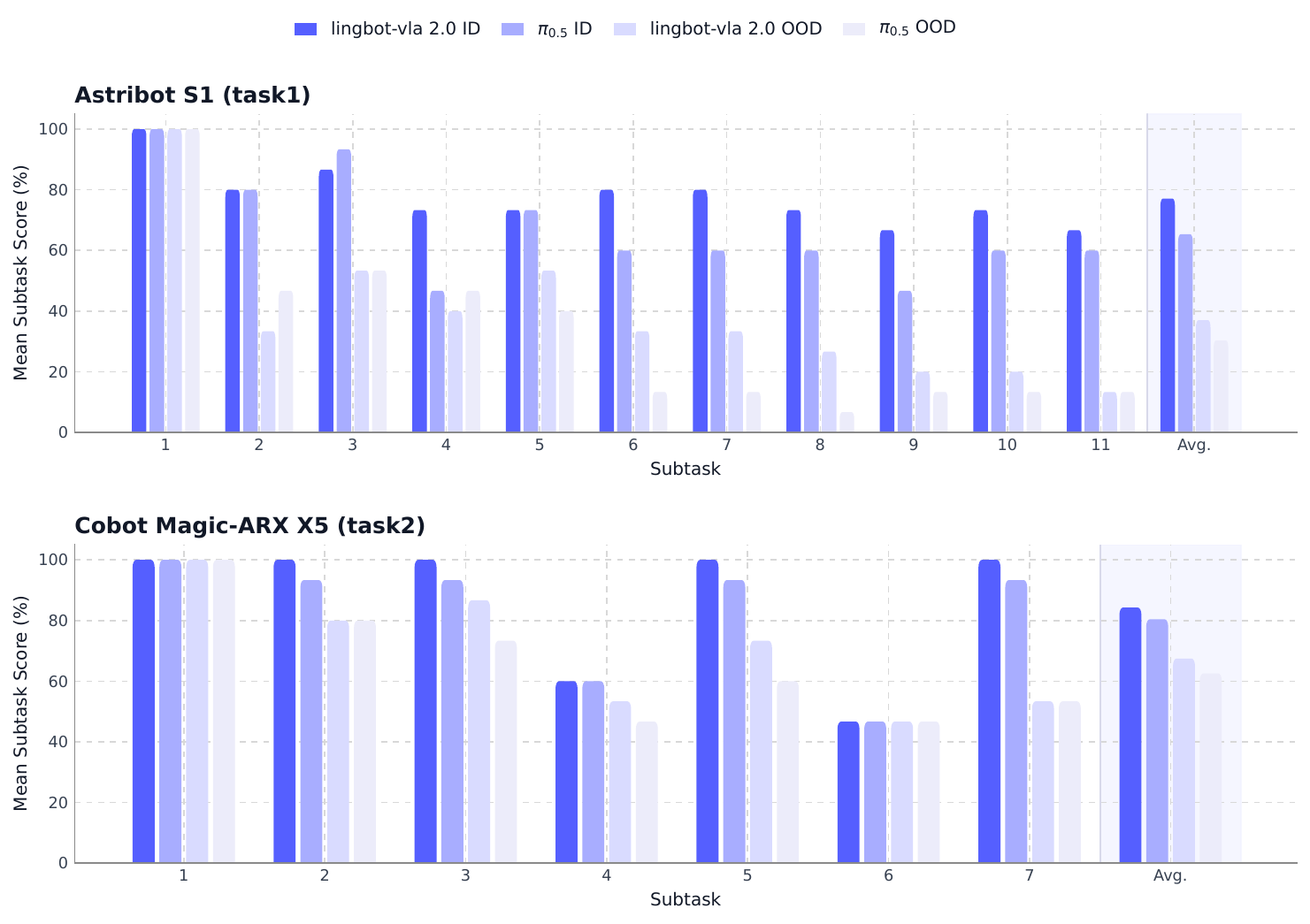}
\caption{Per-subtask performance on the long-horizon mobile manipulation benchmark. Bars report the mean subtask completion score (\%) across 15 trials under both in-domain (ID) and out-of-distribution (OOD) settings.}
\label{fig:bm_subtask_progress_domain_bar}
\end{figure}

\begin{table}[h]
  \centering
  \caption{Long-horizon mobile manipulation benchmark results, reported as progress score / success rate (\%).}
  \label{tab:bm_results}
  \begin{tabular}{lllcc}
  \toprule
  Embodiment & Task & Setting & LingBot-VLA-2.0 & $\pi_{0.5}$ \\
  \midrule
  \multirow{2}{*}{Astribot S1}
  & \multirow{2}{*}{Sort objects into refrigerator}
  & In-domain & 77.1 / 60.0 & 65.3 / 46.7 \\
  & & Out-of-distribution & 37.0 / 13.3 & 30.3 / 6.7 \\
  \midrule
  \multirow{2}{*}{Cobot Magic-ARX X5}
  & \multirow{2}{*}{Stove cleaning}
  & In-domain & 84.3 / 66.7 & 79.9 / 60.0 \\
  & & Out-of-distribution & 67.5 / 40.0 & 62.5 / 33.3 \\
  \bottomrule
\end{tabular}
\end{table}

\begin{figure}[h!]
\centering

\begin{minipage}[t]{0.55\textwidth}
\vspace{0pt}
\centering
\scriptsize
\setlength{\tabcolsep}{3pt}
\renewcommand{\arraystretch}{1.1}
\captionof{table}{Scoring criteria for the long-horizon mobile manipulation experiment tasks.}
\label{tab:bm_scoring_mobile}
\begin{tabular}{p{0.18\linewidth} p{0.78\linewidth}}
\toprule
\textbf{Task} & \textbf{Scoring Criteria} \\
\midrule

Sort objects into refrigerator &
\makecell[l]{
1. Move from the initial position to the kitchen island (6). \\
2. Pick up the drink from the table and place it into the basket (9). \\
3. Pick up the first fruit from the table and place it into the basket (9). \\
4. Pick up the second fruit from the table and place it into the basket (11). \\
5. Pick up and lift the basket (8). \\
6. Move to the front of the refrigerator (10). \\
7. Open the refrigerator door wide enough to place items inside (12). \\
8. Pick up the first fruit and place it into the refrigerator (9). \\
9. Pick up the second fruit and place it into the refrigerator (11). \\
10. Pick up the drink and place it into the refrigerator (9). \\
11. Close the refrigerator door (6).
} \\
\midrule

Stove Cleaning &
\makecell[l]{
1. Move from the initial position to the front of the stove (8). \\
2. Pick up the black pot rack and place it on the left side of the table (16). \\
3. Pick up the sponge (12). \\
4. Wipe off the white foam on the stove using the sponge (18). \\
5. Transfer the sponge and place it on the right side of the stove (16). \\
6. Pick up the black rack and place it back on the stove (16). \\
7. Pick up the black pot and place it on the left burner of the stove (14).
} \\
\bottomrule
\end{tabular}
\end{minipage}
\hfill
\begin{minipage}[t]{0.42\textwidth}
\vspace{1.0cm}
\centering
\includegraphics[width=\linewidth]{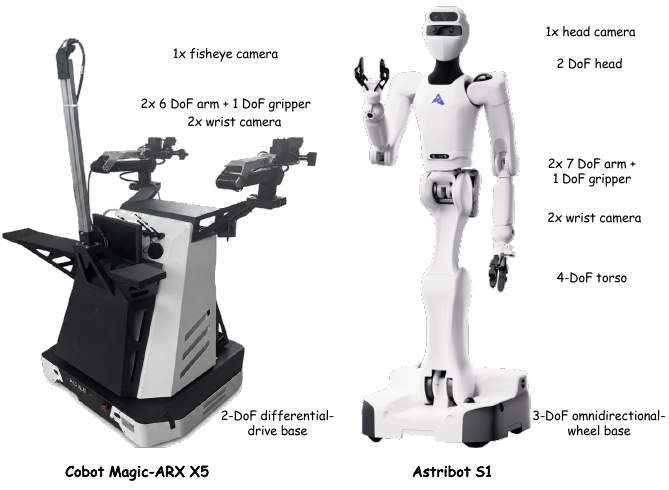}
\captionof{figure}{Illustration of the long-horizon mobile manipulation experiments platform.}
\label{fig:bm_mobile_tasks}
\end{minipage}

\end{figure}

\noindent\textbf{Experiment settings.} We evaluate long-horizon mobile manipulation performance on two robotic embodiments, as illustrated in ~\cref{fig:bm_mobile_tasks}. Each embodiment is associated with one representative task: Astribot S1 is evaluated on the object sorting task, where objects are collected and placed into a refrigerator, while Cobot Magic-ARX X5 is evaluated on the stove cleaning task. The detailed sub-task decomposition and corresponding scoring criteria for each task are listed in ~\cref{tab:bm_scoring_mobile}.
For each task, we evaluate each model under two settings: in-domain (ID) and out-of-distribution (OOD). Each task-setting pair is evaluated with 15 independent trials. In the in-domain setting, both the robot initial position and the manipulated objects are drawn from the training distribution. In the out-of-distribution setting, the robot initial position is perturbed within ±10 cm along the forward, backward, left, and right directions around the nominal initial pose. In addition, for the refrigerator sorting task, the two fruits and the water bottle to be placed into the refrigerator are replaced with unseen object categories. This setting is designed to evaluate the model’s generalization ability under unseen object and position conditions.

\noindent\textbf{Results analysis.} As shown in~\cref{tab:bm_results}, LingBot-VLA-2.0 consistently outperforms $\pi_{0.5}$ across both robotic embodiments and both evaluation settings. In the in-domain setting, LingBot-VLA-2.0 achieves a progress score / success rate of 77.1 / 60.0 on the refrigerator sorting task and 84.3 / 66.7 on the stove cleaning task, improving over $\pi_{0.5}$ by 11.8 / 13.3 and 4.4 / 6.7 points, respectively. These results indicate that LingBot-VLA-2.0 can better execute long-horizon task sequences that require coordinated base movement, object manipulation, and interaction with articulated objects.
Under the OOD setting, both models exhibit performance degradation, reflecting the increased difficulty introduced by perturbed initial robot poses and unseen manipulated objects. Nevertheless, LingBot-VLA-2.0 maintains a clear advantage over $\pi_{0.5}$, achieving 37.0 / 13.3 on the refrigerator-sorting task and 67.5 / 40.0 on the stove-cleaning task. Compared with $\pi_{0.5}$, this corresponds to improvements of 6.7 / 6.6 and 5.0 / 6.7 points, respectively. The performance gap is especially meaningful in terms of success rate, suggesting that LingBot-VLA-2.0 is more robust in completing full task trajectories rather than only making partial progress.
The refrigerator sorting task shows a larger drop from the in-domain to the out-of-distribution setting than the stove cleaning task. This is expected because the out-of-distribution refrigerator setting simultaneously changes both the initial robot pose and the manipulated object categories, requiring stronger object-level generalization and more precise long-horizon recovery. In contrast, the stove cleaning task mainly evaluates robustness to initial pose perturbations while preserving the task objects and scene structure. Overall, the results demonstrate that LingBot-VLA-2.0 achieves stronger long-horizon mobile manipulation capability and better generalization than $\pi_{0.5}$ across both embodiments.

\section{Ablation Studies}

\subsection{Action Space}
\begin{figure*}[h]
    \centering
    \includegraphics[width=\textwidth, trim=0 0 0 50pt, clip]{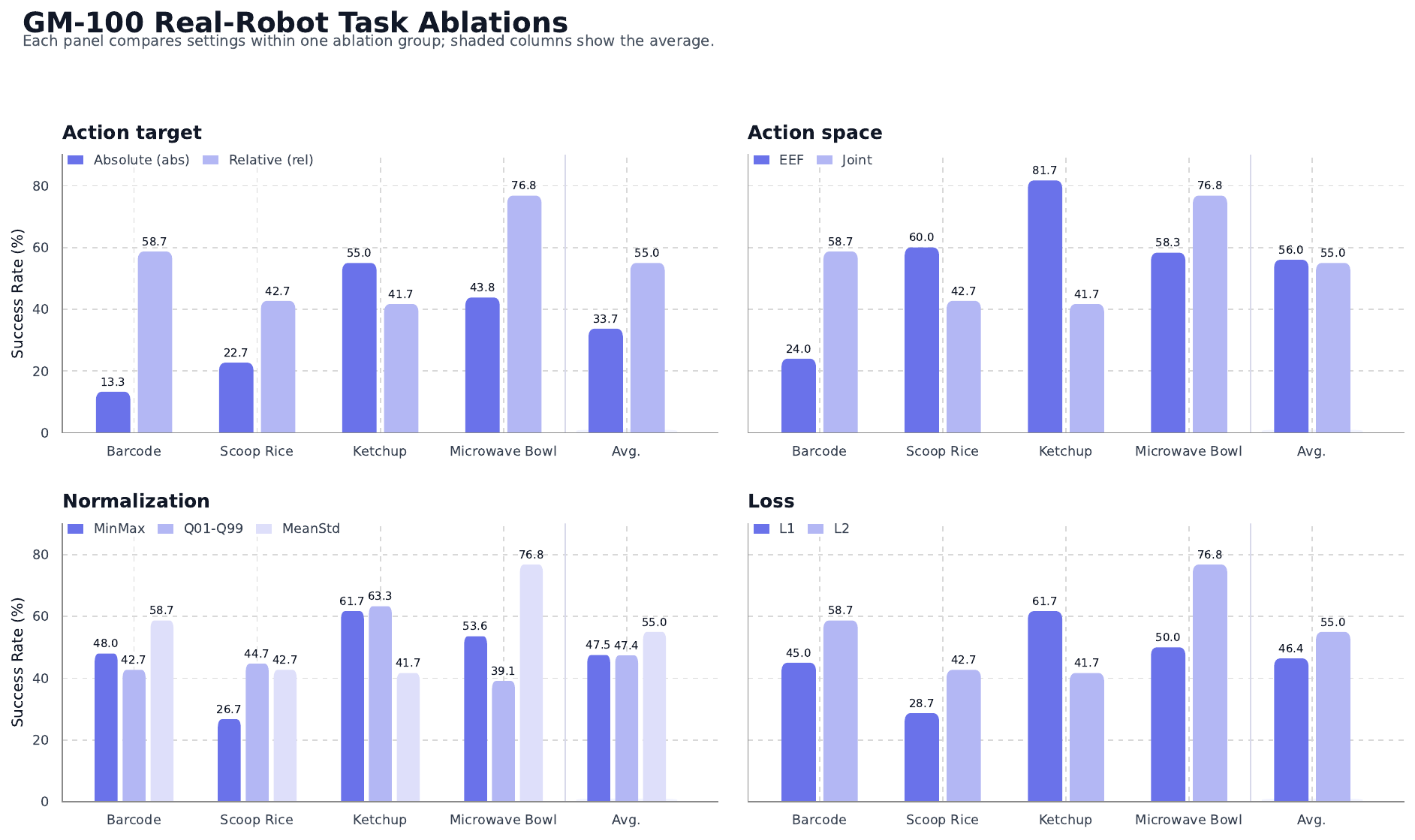}
    \caption{Results on the four GM-100 real-robot tasks.}
    \label{fig:gm100}
\end{figure*}

\begin{figure*}[h!]
    \centering
    \includegraphics[width=\textwidth]{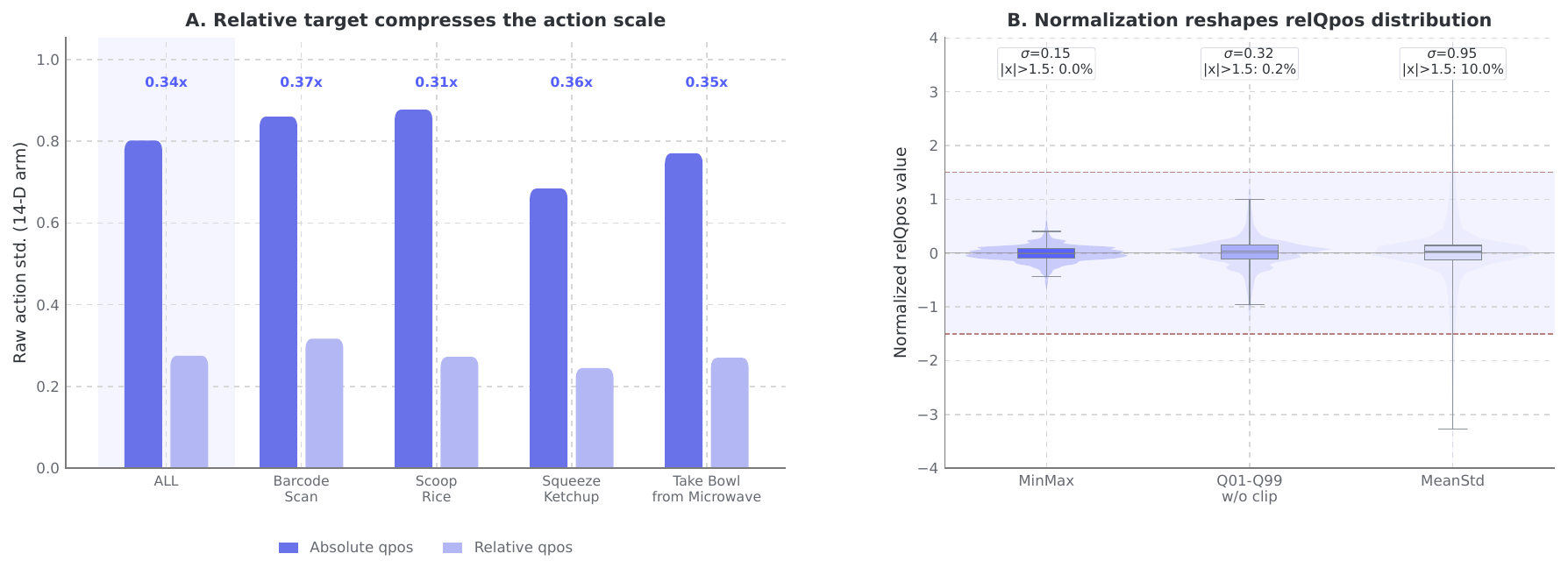}
    \caption{Action targets and normalization statistics for the four GM-100 tasks. Left: \texttt{relQpos} substantially reduces the action scale compared with \texttt{absQpos}. Right: normalization changes the effective dynamic range of MeanStd-normalized \texttt{relQpos}; Q01--Q99 is shown in the actual unclipped setting.}
    \label{fig:action_target_norm_stats}
\end{figure*}

\noindent\textbf{Action target.}
For the action target, relative joint actions substantially outperform absolute joint actions, improving the average success rate from $33.7$ to $55.0$. This is supported by the action statistics in \cref{fig:action_target_norm_stats}A. Across the four tasks, the standard deviation of \texttt{relQpos} is only $31\%$--$37\%$ of that of \texttt{absQpos}; on the pooled distribution, the action standard deviation decreases from about $0.80$ for \texttt{absQpos} to $0.28$ for \texttt{relQpos}. Therefore, relative actions convert the prediction target from global joint-configuration regression into local motion regression, producing targets that are more centered and have lower variance.

\begin{figure*}[h!]
    \centering
    \includegraphics[width=\textwidth]{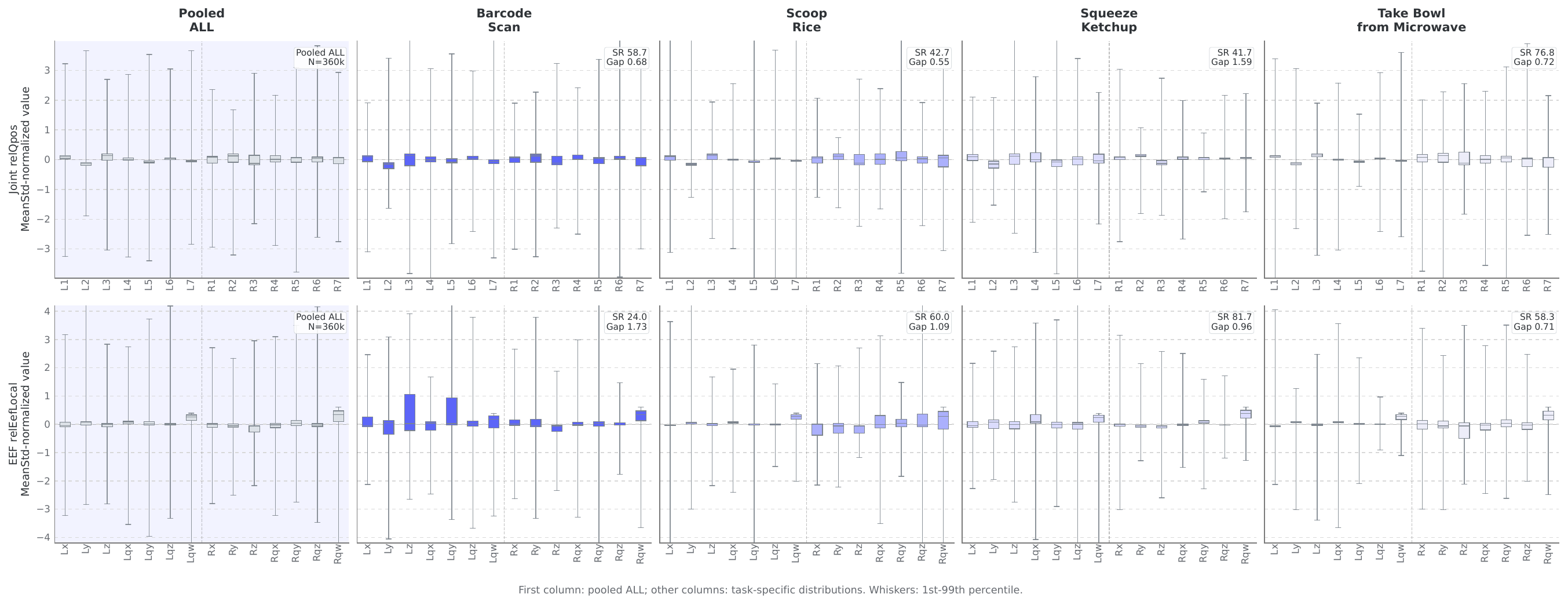}
    \caption{Per-dimension MeanStd-normalized action distributions for joint and EEF action spaces. The first column shows the pooled distribution over all four tasks, while the remaining columns show task-specific distributions. The gap measures task-to-pooled distribution alignment based on per-dimension median and IQR differences.}
    \label{fig:action_space_boxalign}
\end{figure*}

\noindent\textbf{Action space.}
For the action space, EEF and joint actions obtain similar average success rates, $56.0$ and $55.0$, respectively, but show different task-level preferences. To analyze this, \Cref{fig:action_space_boxalign} compares the per-dimension MeanStd-normalized action distributions of each task against the pooled distribution over all four tasks. We additionally report a distribution-alignment gap, computed from the per-dimension median and IQR differences between each task and the pooled distribution; a smaller gap indicates that the task-specific action distribution is closer to the overall distribution.

The results show that distribution alignment explains part, but not all, of the action-space behavior. For Barcode Scan, joint actions are much closer to the pooled distribution than EEF actions, with a gap of $0.68$ versus $1.73$, consistent with the large performance advantage of joint actions ($58.7$ vs. $24.0$). For Squeeze Ketchup, the opposite trend appears: EEF actions have a smaller gap than joint actions ($0.96$ vs. $1.59$) and achieve much higher success ($81.7$ vs. $41.7$), suggesting that contact-rich endpoint motions are better represented in Cartesian EEF space. For Scoop Rice, however, EEF actions outperform joint actions despite a larger distribution gap, indicating that Cartesian motion regularity can outweigh pure distribution alignment. For Take Bowl from Microwave, the two gaps are nearly identical ($0.71$--$0.72$), while joint actions still perform better, suggesting that posture, reachability, and configuration-dependent constraints are more important than marginal action statistics. Overall, the best action space depends on both distributional alignment and the physical structure of the task.

\noindent\textbf{Normalization.}
For normalization, \Cref{fig:action_target_norm_stats}B shows the normalized \texttt{relQpos} distributions under the three schemes. MinMax compresses most samples into a narrow range: its normalized standard deviation is only $0.15$, and the 1st--99th percentile interval is approximately $[-0.42, 0.40]$. This strong compression reduces the effective resolution of the action targets, which explains its lower average success rate of $47.5$.

Q01--Q99 increases the normalized standard deviation to $0.32$ and expands the central mass to roughly $[-0.96, 1.01]$. Importantly, the actual implementation used here is not clipped; only about $0.2\%$ of values fall outside $[-1.5, 1.5]$. Thus, its weakness is not information loss from clipping, but rather that the normalized targets remain much more compressed than under MeanStd. This explains why Q01--Q99 improves the central dynamic range over MinMax but still achieves a similar average success rate of $47.4$.

MeanStd provides the largest effective dynamic range, with normalized standard deviation $0.95$ and a 1st--99th percentile interval of approximately $[-3.22, 3.30]$. Around $10.0\%$ of samples lie outside $|x|>1.5$, reflecting the long-tailed structure of the relative action distribution rather than clipping. By preserving these larger corrective motions, MeanStd achieves the best average success rate of $55.0$.

\noindent\textbf{Loss function.}
For the loss function, L2 achieves the best average performance, improving over L1 from $46.4$ to $55.0$. Since most \texttt{relQpos} targets are small continuous corrections around zero, L2 better fits the high-density region of the action distribution and encourages precise regression. L1 performs better on Squeeze Ketchup, consistent with its robustness to contact-rich or heavy-tailed motions, but underperforms on the other three tasks.

\subsection{Perceptual Results of Dual-query Distillation}
\begin{figure}[t]
\centering
\includegraphics[width=\linewidth]{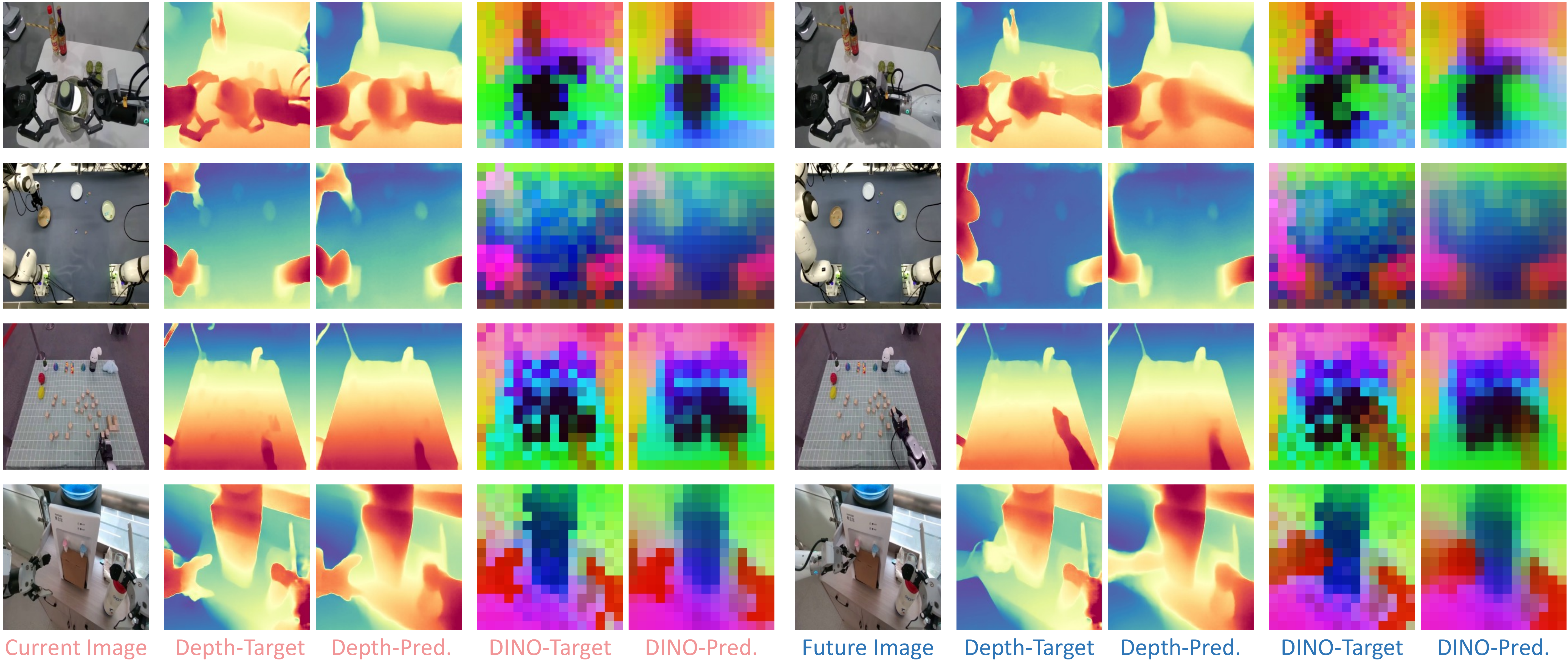}
\caption{Causal perception results of \method under visual distillation. \textbf{Left}:  depth/DINO-Video-PCA ground truth and prediction for current image. \textbf{Right}: depth/DINO-Video-PCA ground truth and prediction for future image.}
\label{fig:vis_distillation}
\end{figure}

We adopt LingBot-Depth and DINO-Video as the visual teacher models for distillation training. Despite perceptual results not being essential for action generation, integrating specific task queries allows the VLA to obtain visual perceptual outcomes during causal inference. As shown in~\cref{fig:vis_distillation}, we present the causal perception results of \method, which validate the effectiveness of our model in distilling semantic priors and geometric cues.

\section{Conclusion}\label{sec:conclusion}
\method narrows the gap between VLA foundation models and real-world robotic deployment by improving generalization, expanding whole-body action modeling, and strengthening temporal reasoning through predictive dynamics. With large-scale robot and egocentric pretraining data, \method achieves improved performance on the GM-100 benchmark and demonstrates strong long-horizon mobile manipulation across multiple robotic platforms.

\noindent\textbf{Acknowledgment.}
We thank Fangyi Xu, Ningyuan Huang, Haotian Liu, Jialiang Zheng, Liping Zhang, Weilun Yao, Shuai Wang, Linyu Su, Fan Fan, Bo Jiang, Jingmei Zhao, Shuai Zhou, Yinghao Xu and Haiwei Liang for help with data, evaluation experiments, training infrastructure, robot hardware, and robot software.
We also gratefully acknowledge Ant Digital Technologies' Phecda Laboratory and Genrobot.ai Co., Ltd. for providing the egocentric data.

{
\small
\bibliographystyle{plain}
\bibliography{ref.bib}

\begin{thebibliography}{10}

\bibitem{cosmos3}
Niket Agarwal, Arslan Ali, Jon Allen, Martin Antolini, Adeline Aubame, Alisson Azzolini, Junjie Bai, Maciej Bala, Yogesh Balaji, Josh Bapst, et~al.
\newblock Cosmos 3: Omnimodal world models for physical ai.
\newblock {\em arXiv preprint arXiv:2606.02800}, 2026.

\bibitem{vjepa2}
Mahmoud Assran, Adrien Bardes, David Fan, Quentin Garrido, Russell Howes, Mojtaba Komeili, Matthew Muckley, Ammar Rizvi, Claire Roberts, Koustuv Sinha, Artem Zholus, Sergio Arnaud, Abha Gejji, Ada Martin, Francois Robert~Hogan, Daniel Dugas, Piotr Bojanowski, Vasil Khalidov, Patrick Labatut, Francisco Massa, Marc Szafraniec, Kapil Krishnakumar, Yong Li, Xiaodong Ma, Sarath Chandar, Franziska Meier, Yann LeCun, Michael Rabbat, and Nicolas Ballas.
\newblock V-jepa~2: Self-supervised video models enable understanding, prediction and planning.
\newblock {\em arXiv preprint arXiv:2506.09985}, 2025.

\bibitem{hex}
Shuanghao Bai, Meng Li, Xinyuan Lv, Jiawei Wang, Xinhua Wang, Fei Liao, Chengkai Hou, Langzhe Gu, Wanqi Zhou, Kun Wu, Ziluo Ding, Zhiyuan Xu, Lei Sun, Shanghang Zhang, Zhengping Che, Jian Tang, and Badong Chen.
\newblock {HEX}: Humanoid-aligned experts for cross-embodiment whole-body manipulation.
\newblock {\em arXiv preprint arXiv:2604.07993}, 2026.

\bibitem{gr00tN1}
Johan Bjorck, Fernando Casta{\~n}eda, Nikita Cherniadev, Xingye Da, Runyu Ding, Linxi Fan, Yu~Fang, Dieter Fox, Fengyuan Hu, Spencer Huang, et~al.
\newblock {GR00T N1}: {A}n open foundation model for generalist humanoid robots.
\newblock {\em arXiv preprint arXiv:2503.14734}, 2025.

\bibitem{pi_0d5}
Kevin Black, Noah Brown, James Darpinian, Karan Dhabalia, Danny Driess, Adnan Esmail, Michael~Robert Equi, Chelsea Finn, Niccolo Fusai, Manuel~Y. Galliker, Dibya Ghosh, Lachy Groom, Karol Hausman, brian ichter, Szymon Jakubczak, Tim Jones, Liyiming Ke, Devin LeBlanc, Sergey Levine, Adrian Li-Bell, Mohith Mothukuri, Suraj Nair, Karl Pertsch, Allen~Z. Ren, Lucy~Xiaoyang Shi, Laura Smith, Jost~Tobias Springenberg, Kyle Stachowicz, James Tanner, Quan Vuong, Homer Walke, Anna Walling, Haohuan Wang, Lili Yu, and Ury Zhilinsky.
\newblock $\pi_{0.5}$: A vision-language-action model with open-world generalization.
\newblock In {\em Conference on Robot Learning}, 2025.

\bibitem{pi_0}
Kevin Black, Noah Brown, Danny Driess, Adnan Esmail, Michael Equi, Chelsea Finn, Niccolo Fusai, Lachy Groom, Karol Hausman, Brian Ichter, Szymon Jakubczak, Tim Jones, Liyiming Ke, Sergey Levine, Adrian Li-Bell, Mohith Mothukuri, Suraj Nair, Karl Pertsch, Lucy~Xiaoyang Shi, James Tanner, Quan Vuong, Anna Walling, Haohuan Wang, and Ury Zhilinsky.
\newblock $\pi_{0}$: A vision-language-action flow model for general robot control.
\newblock In {\em Proceedings of Robotics: Science and Systems}, 2025.

\bibitem{cai2026xiaomi}
Rui Cai, Jun Guo, Xinze He, Piaopiao Jin, Jie Li, Bingxuan Lin, Futeng Liu, Wei Liu, Fei Ma, Kun Ma, Feng Qiu, Heng Qu, Yifei Su, Qiao Sun, Dong Wang, Donghao Wang, Yunhong Wang, Rujie Wu, Diyun Xiang, Yu~Yang, Hangjun Ye, Yuan Zhang, and Quanyun Zhou.
\newblock Xiaomi-robotics-0: An open-sourced vision-language-action model with real-time execution.
\newblock {\em arXiv preprint arXiv:2602.12684}, 2026.

\bibitem{gr3}
Chilam Cheang, Sijin Chen, Zhongren Cui, Yingdong Hu, Liqun Huang, Tao Kong, Hang Li, Yifeng Li, Yuxiao Liu, Xiao Ma, et~al.
\newblock {GR-3} technical report.
\newblock {\em arXiv preprint arXiv:2507.15493}, 2025.

\bibitem{lawam}
Jialei Chen, Kai Wang, Kang Chen, Shuaihang Chen, Feng Gao, Wenhao Tang, Zhiyuan Li, Weilin Liu, Zhuyu Yao, Boxun Li, et~al.
\newblock Lawam: Latent world action models for efficient dynamics-aware robot policies.
\newblock {\em arXiv preprint arXiv:2606.15768}, 2026.

\bibitem{abotm05}
Ronghan Chen, Yandan Yang, Zuojin Tang, Dongjie Huo, Tong Lin, Haoning Wu, Haoyun Liu, Yuzhi Chen, Lulu Zheng, Botai Yuan, Tianlun Li, Mingxin Wang, Dekang Qi, Bin Hu, Wei Mei, Yuze Xuan, Haolong Yang, Yanqing Zhu, Mu~Xu, Zhiheng Ma, and Xinyuan Chang.
\newblock Abot-m0.5: Unified mobility-and-manipulation world action model.
\newblock {\em arXiv preprint arXiv:2607.00678}, 2026.

\bibitem{dexWorldModel}
{DexForce AI Team of Physical AI}.
\newblock Dexworldmodel: Causal latent world modeling towards automated learning of embodied tasks, 2026.
\newblock Technical report.

\bibitem{himoevla}
Zhiying Du, Bei Liu, Yaobo Liang, Yichao Shen, Haidong Cao, Xiangyu Zheng, Zhiyuan Feng, Zuxuan Wu, Jiaolong Yang, and Yu-Gang Jiang.
\newblock {HiMoE-VLA}: Hierarchical mixture-of-experts for generalist vision-language-action policies.
\newblock {\em arXiv preprint arXiv:2512.05693}, 2025.

\bibitem{g0_5}
{Galaxea Team}.
\newblock Galaxea g0.5 technical report, 2026.
\newblock Technical report.

\bibitem{galaxeaG0}
Tao Jiang, Tianyuan Yuan, Yicheng Liu, Chenhao Lu, Jianning Cui, Xiao Liu, Shuiqi Cheng, Jiyang Gao, Huazhe Xu, and Hang Zhao.
\newblock Galaxea open-world dataset and {G}0 dual-system vla model.
\newblock {\em arXiv preprint arXiv:2509.00576}, 2025.

\bibitem{rldx1}
Dongyoung Kim, Huiwon Jang, Myungkyu Koo, Suhyeok Jang, Taeyoung Kim, Beomjun Kim, Byungjun Yoon, Changsung Jang, Daewon Choi, Dongsu Han, Donguk Lee, Heeseung Kwon, Hojin Jeon, Jaehyun Kang, Jaekyoung Bae, Jihyuk Lee, Jimin Lee, John Won, Joonwoo Ahn, Junhyeong Park, Junyoung Sung, Kyungmin Lee, Minseong Han, Minsung Yoon, Sejune Joo, Seonil Son, Seungcheol Park, Seunggeun Cho, Seungjun Moon, Seungku Kim, Yonghoon Dong, Yongjin Cho, Youngchan Kim, et~al.
\newblock {RLDX}-1 technical report.
\newblock {\em arXiv preprint arXiv:2605.03269}, 2026.

\bibitem{openvla}
Moo~Jin Kim, Karl Pertsch, Siddharth Karamcheti, Ted Xiao, Ashwin Balakrishna, Suraj Nair, Rafael Rafailov, Ethan~P Foster, Pannag~R Sanketi, Quan Vuong, et~al.
\newblock {OpenVLA}: An open-source vision-language-action model.
\newblock In {\em Conference on Robot Learning}, 2025.

\bibitem{forcevla2}
Yang Li, Zhaxizhuoma, Hongru Jiang, Junjie Xia, Hongquan Zhang, Jinda Du, Yunsong Zhou, Jia Zeng, Ce~Hao, Jieji Ren, Qiaojun Yu, Cewu Lu, Yu~Qiao, and Jiangmiao Pang.
\newblock {ForceVLA2}: Unleashing hybrid force-position control with force awareness for contact-rich manipulation.
\newblock {\em arXiv preprint arXiv:2603.15169}, 2026.

\bibitem{holoBrain0}
Xuewu Lin, Tianwei Lin, Yun Du, Hongyu Xie, Yiwei Jin, Jiawei Li, Shijie Wu, Qingze Wang, Mengdi Li, Mengao Zhao, Ziang Li, Chaodong Huang, Hongzhe Bi, Lichao Huang, and Zhizhong Su.
\newblock Holobrain-0 technical report.
\newblock {\em arXiv preprint arXiv:2602.12062}, 2026.

\bibitem{liu2024deepseek}
Aixin Liu, Bei Feng, Bing Xue, Bingxuan Wang, Bochao Wu, Chengda Lu, Chenggang Zhao, Chengqi Deng, Chenyu Zhang, Chong Ruan, et~al.
\newblock Deepseek-v3 technical report.
\newblock {\em arXiv preprint arXiv:2412.19437}, 2024.

\bibitem{beingH0}
Hao Luo, Yicheng Feng, Wanpeng Zhang, Sipeng Zheng, Ye~Wang, Haoqi Yuan, Jiazheng Liu, Chaoyi Xu, Qin Jin, and Zongqing Lu.
\newblock Being-h0: Vision-language-action pretraining from large-scale human videos.
\newblock {\em arXiv preprint arXiv:2507.15597}, 2025.

\bibitem{beingH05}
Hao Luo, Ye~Wang, Wanpeng Zhang, Sipeng Zheng, Ziheng Xi, Chaoyi Xu, Haiweng Xu, Haoqi Yuan, Chi Zhang, Yiqing Wang, Yicheng Feng, and Zongqing Lu.
\newblock Being-h0.5: Scaling human-centric robot learning for cross-embodiment generalization.
\newblock {\em arXiv preprint arXiv:2601.12993}, 2026.

\bibitem{beingH07}
Hao Luo, Wanpeng Zhang, Yicheng Feng, Sipeng Zheng, Haiweng Xu, Chaoyi Xu, Ziheng Xi, Yuhui Fu, and Zongqing Lu.
\newblock Being-h0.7: A latent world-action model from egocentric videos.
\newblock {\em arXiv preprint arXiv:2605.00078}, 2026.

\bibitem{lda1b}
Jiangran Lyu, Kai Liu, Xuheng Zhang, Haoran Liao, Yusen Feng, Wenxuan Zhu, Tingrui Shen, Jiayi Chen, Jiazhao Zhang, Yifei Dong, Wenbo Cui, Senmao Qi, Shuo Wang, Yixin Zheng, Mi~Yan, Xuesong Shi, Haoran Li, Dongbin Zhao, Ming-Yu Liu, Zhizheng Zhang, Li~Yi, Yizhou Wang, and He~Wang.
\newblock {LDA}-1b: Scaling latent dynamics action model via universal embodied data ingestion.
\newblock {\em arXiv preprint arXiv:2602.12215}, 2026.

\bibitem{larybench}
Dujun Nie, Fengjiao Chen, Qi~Lv, Jun Kuang, Xiaoyu Li, Xuezhi Cao, and Xunliang Cai.
\newblock Lary: A latent action representation yielding benchmark for generalizable vision-to-action alignment.
\newblock {\em arXiv preprint arXiv:2604.11689}, 2026.

\bibitem{qwen3.6-27b}
{Qwen Team}.
\newblock {Qwen3.6-27B}: Flagship-level coding in a {27B} dense model, April 2026.

\bibitem{dinov3}
Oriane Sim{\'e}oni, Huy~V. Vo, Maximilian Seitzer, Federico Baldassarre, Maxime Oquab, Cijo Jose, Vasil Khalidov, Marc Szafraniec, Seungeun Yi, Micha{\"e}l Ramamonjisoa, Francisco Massa, Daniel Haziza, Luca Wehrstedt, Jianyuan Wang, Timoth{\'e}e Darcet, Th{\'e}o Moutakanni, Leonel Sentana, Claire Roberts, Andrea Vedaldi, Jamie Tolan, John Brandt, Camille Couprie, Julien Mairal, Herv{\'e} J{\'e}gou, Patrick Labatut, and Piotr Bojanowski.
\newblock {DINOv3}, 2025.

\bibitem{wog}
Yue Su, Sijin Chen, Haixin Shi, Mingyu Liu, Zhengshen Zhang, Ningyuan Huang, Weiheng Zhong, Zhengbang Zhu, Yuxiao Liu, and Xihui Liu.
\newblock World guidance: World modeling in condition space for action generation.
\newblock In {\em ICML}, 2026.

\bibitem{lingbotdepth}
Bin Tan, Changjian Sun, Xiage Qin, Hanat Adai, Zelin Fu, Tianxiang Zhou, Han Zhang, Yinghao Xu, Xing Zhu, Yujun~Shen Shen, and Nan Xue.
\newblock Masked depth modeling for spatial perception.
\newblock \url{https://technology.robbyant.com/lingbot-depth}, 2026.

\bibitem{modevla}
Tutian Tang, Xingyu Ji, Wanli Xing, Ce~Hao, Wenqiang Xu, Lin Shao, Cewu Lu, Qiaojun Yu, Jiangmiao Pang, and Kaifeng Zhang.
\newblock Towards human-like manipulation through {RL}-augmented teleoperation and mixture-of-dexterous-experts {VLA}.
\newblock {\em arXiv preprint arXiv:2603.08122}, 2026.

\bibitem{gemini_robotics}
Gemini~Robotics Team, Saminda Abeyruwan, Joshua Ainslie, Jean-Baptiste Alayrac, Montserrat~Gonzalez Arenas, Travis Armstrong, Ashwin Balakrishna, Robert Baruch, Maria Bauza, Michiel Blokzijl, et~al.
\newblock Gemini {R}obotics: Bringing {AI} into the physical world.
\newblock {\em arXiv preprint arXiv:2503.20020}, 2025.

\bibitem{generalist2026gen1}
Generalist Team.
\newblock Gen-1: Scaling embodied foundation models to mastery.
\newblock {\em Generalist AI Blog}, 2026.
\newblock https://generalistai.com/blog/gen-1.

\bibitem{gr00t_n1_6}
NVIDIA~GEAR Team.
\newblock {GR00T N1.6}: An improved open foundation model for generalist humanoid robots.
\newblock \url{https://research.nvidia.com/labs/gear/gr00t-n1_6/}, 2025.

\bibitem{qwenrobotmanip}
Qwen Team.
\newblock Qwen-robotmanip technical report: Alignment unlocks scale for robotic manipulation foundation models.
\newblock 2026.

\bibitem{qwenVLA}
Qiuyue Wang, Mingsheng Li, Jian Guan, Jinhui Ye, Sicheng Xie, Yitao Liu, Junhao Chen, Zhixuan Liang, Jie Zhang, Xintong Hu, Xuhong Huang, Pei Lin, Junyang Lin, Dayiheng Liu, Shuai Bai, Jingren Zhou, Jiazhao Zhang, Haoqi Yuan, Gengze Zhou, Hang Yin, Ye~Wang, Yiyang Huang, Zixing Lei, Wujian Peng, Delin Chen, et~al.
\newblock {Qwen}-{VLA}: Unifying vision-language-action modeling across tasks, environments, and robot embodiments.
\newblock {\em arXiv preprint arXiv:2605.30280}, 2026.

\bibitem{wang2025vision}
Yunnan Wang, Fan Lu, Kecheng Zheng, Ziyuan Huang, Ziqiang Li, Wenjun Zeng, and Xin Jin.
\newblock Vision-centric activation and coordination for multimodal large language models.
\newblock {\em arXiv preprint arXiv:2510.14349}, 2025.

\bibitem{gm100}
Ziyu Wang, Chenyuan Liu, Yushun Xiang, Runhao Zhang, Qingbo Hao, Hongliang Lu, Houyu Chen, Zhizhong Feng, Kaiyue Zheng, Dehao Ye, Xianchao Zeng, Xinyu Zhou, Boran Wen, Jiaxin Li, Mingyu Zhang, Kecheng Zheng, Qian Zhu, Ran Cheng, and Yong-Lu Li.
\newblock {The Great March 100}: 100 detail-oriented tasks for evaluating embodied ai agents, 2026.

\bibitem{videorope}
Xilin Wei, Xiaoran Liu, Yuhang Zang, Xiaoyi Dong, Pan Zhang, Yuhang Cao, Jian Tong, Haodong Duan, Qipeng Guo, Jiaqi Wang, et~al.
\newblock Videorope: What makes for good video rotary position embedding?
\newblock In {\em Int. Conf. Mach. Learn.}, 2025.

\bibitem{lingbotvla}
Wei Wu, Fan Lu, Yunnan Wang, Shuai Yang, Shi Liu, Fangjing Wang, Shuailei Ma, He~Sun, Yong Wang, Zhenqi Qiu, Houlong Xiong, Ziyu Wang, Shuai Zhou, Yiyu Ren, Kejia Zhang, Hui Yu, Jingmei Zhao, Qian Zhu, Ran Cheng, Yong-Lu Li, Yongtao Huang, Xing Zhu, Yujun Shen, and Kecheng Zheng.
\newblock A pragmatic vla foundation model.
\newblock {\em arXiv preprint arXiv:2601.18692v1}, 2026.

\bibitem{hyembodied05x}
Tencent~Robotics X and HY~Vision Team.
\newblock Hy-embodied-0.5-x: An enhanced embodied foundation model for real-world agents.
\newblock 2026.

\bibitem{omnistream}
Yibin Yan, Jilan Xu, Shangzhe Di, Haoning Wu, and Weidi Xie.
\newblock Omnistream: Mastering perception, reconstruction and action in continuous streams.
\newblock {\em arXiv preprint arXiv:2603.12265}, 2026.

\bibitem{magma}
Jianwei Yang, Reuben Tan, Qianhui Wu, Ruijie Zheng, Baolin Peng, Yongyuan Liang, Yu~Gu, Mu~Cai, Seonghyeon Ye, Joel Jang, et~al.
\newblock Magma: A foundation model for multimodal {AI} agents.
\newblock In {\em IEEE Conf. Comput. Vis. Pattern Recog.}, pages 14203--14214, 2025.

\bibitem{abotm0}
Yandan Yang, Shuang Zeng, Tong Lin, Xinyuan Chang, Dekang Qi, Junjin Xiao, Haoyun Liu, Ronghan Chen, Yuzhi Chen, Dongjie Huo, et~al.
\newblock Abot-m0: Vla foundation model for robotic manipulation with action manifold learning.
\newblock {\em arXiv preprint arXiv:2602.11236}, 2026.

\bibitem{starVLAalpha}
Jinhui Ye, Ning Gao, Senqiao Yang, Jinliang Zheng, Zixuan Wang, Yuxin Chen, Pengguang Chen, Yilun Chen, Shu Liu, and Jiaya Jia.
\newblock Starvla-$\\alpha$: Reducing complexity in vision-language-action systems.
\newblock {\em arXiv preprint arXiv:2604.11757}, 2026.

\bibitem{samoevla}
Zihan You, Hongwei Liu, Chenxu Dang, Zhe Wang, Sining Ang, Aoqi Wang, and Yan Wang.
\newblock {SAMoE-VLA}: A scene adaptive mixture-of-experts vision-language-action model for autonomous driving.
\newblock {\em arXiv preprint arXiv:2603.08113}, 2026.

\bibitem{forcevla}
Jiawen Yu, Hairuo Liu, Qiaojun Yu, Jieji Ren, Ce~Hao, Haitong Ding, Guangyu Huang, Guofan Huang, Yan Song, Panpan Cai, Cewu Lu, and Wenqiang Zhang.
\newblock {ForceVLA}: Enhancing {VLA} models with a force-aware {MoE} for contact-rich manipulation.
\newblock {\em arXiv preprint arXiv:2505.22159}, 2025.

\bibitem{wallOSS05}
Ryan Yu, Pushi Zhang, Starrick Liu, Brae Liu, Miracle Kang, Shalfun Li, Lights Shi, Ellie Ma, Ping Yang, Chris Pan, Jerry Chen, Dongxiu Liu, Rain Sun, Miles Guo, Byron Zhang, Hugo Zhou, Zach Xu, Vincent Chen, Harrison Huang, James Wang, Dance Kuzi, Andy Zhai, Hang Su, Roy Gan, Lucy Liang, Hao Wang, and Qian Wang.
\newblock Wall-oss-0.5 technical report.
\newblock {\em arXiv preprint arXiv:2605.30877}, 2026.

\bibitem{walloss}
Andy Zhai, Brae Liu, Bruno Fang, Chalse Cai, Ellie Ma, Ethan Yin, Hao Wang, Hugo Zhou, James Wang, Lights Shi, et~al.
\newblock Igniting {VLMs} toward the embodied space.
\newblock {\em arXiv preprint arXiv:2509.11766}, 2025.

\bibitem{atomicvla}
Likui Zhang, Tao Tang, Zhihao Zhan, Xiuwei Chen, Zisheng Chen, Jianhua Han, Jiangtong Zhu, Pei Xu, Hang Xu, Hefeng Wu, Liang Lin, and Xiaodan Liang.
\newblock {AtomicVLA}: Unlocking the potential of atomic skill learning in robots.
\newblock {\em arXiv preprint arXiv:2603.07648}, 2026.

\bibitem{gem}
Ruowen Zhao, Bangguo Li, Zuyan Liu, Yinan Liang, Junliang Ye, Fangfu Liu, Diankun Wu, Zhengyi Wang, Xumin Yu, Yongming Rao, Han Hu, and Jun Zhu.
\newblock {GEM}: Generative supervision helps embodied intelligence.
\newblock {\em arXiv preprint arXiv:2605.28548}, 2026.

\end{thebibliography}
}

\end{document}